\documentclass[11pt]{article}

\usepackage[final]{acl}
\usepackage{times}
\usepackage{latexsym}
\usepackage{multirow}
\usepackage{tablefootnote}
\usepackage[most]{tcolorbox}
\usepackage{tablefootnote}
\usepackage{wrapfig}
\usepackage{amssymb, amsmath}
\usepackage{pifont}
\newcommand{\xmark}{\ding{55}}%
\usepackage{times}
\usepackage{latexsym}
\usepackage[most]{tcolorbox}

\usepackage[T1]{fontenc}
\usepackage[utf8]{inputenc}
\usepackage{hyperref}
\usepackage{microtype}

\usepackage{inconsolata}

\usepackage{booktabs}
\usepackage{multirow}
\usepackage{newtxmath}
\usepackage[T1]{fontenc}

\usepackage[utf8]{inputenc}

\usepackage{microtype}

\usepackage{inconsolata}

\usepackage{graphicx}

\title{EDUMATH: Generating Standards-aligned Educational Math Word Problems}

\author{
    Bryan R. Christ \newline
    Penelope Molitz  \newline
    Beau LeBlond  \\ \hspace{.4cm}
    \textbf{Zachary Gottesman} \hspace{.4cm} \newline \indent
    \textbf{Jonathan Kropko} \newline \hspace{.4cm}
    \textbf{Thomas Hartvigsen} \\
    University of Virginia \\
    \small{
    \textbf{Correspondence:} \href{mailto:brc4cb@virginia.edu}{brc4cb@virginia.edu}
  }
}

\begin{document}
\maketitle
\begin{abstract}
Math word problems (MWPs) are critical K-12 educational tools, and customizing them to students' interests and ability levels can enhance learning. However, teachers struggle to find time to customize MWPs for students given large class sizes and increasing burnout. We propose that LLMs can support math education by generating MWPs customized to student interests and math education standards. We use a joint human expert-LLM judge approach to evaluate over 11,000 MWPs generated by open and closed LLMs and develop the first teacher-annotated dataset for standards-aligned educational MWP generation. We show the value of our data by using it to train a 12B open model that matches the performance of larger and more capable open models. We also use our teacher-annotated data to train a text classifier that enables a 30B open LLM to outperform existing closed baselines without any training. Next, we show our models' MWPs are more similar to human-written MWPs than those from existing models. We conclude by conducting the first study of customized LLM-generated MWPs with grade school students, finding they perform similarly on our models' MWPs relative to human-written MWPs but consistently prefer our customized MWPs. \footnote{Our code, models, and data can be found at \href{https://github.com/bryanchrist/EDUMATH}{https://github.com/bryanchrist/EDUMATH}}

\end{abstract}

\section{Introduction}
Math word problems (MWPs) are natural language math questions paired with numerical solutions, and are critical elements of K-12 math education because they help verify mastery of a math concept \cite{daroczy_word_2015, article, schwartz_why_2023, verschaffel_word_2020}. In all K-12 grades, personalizing MWPs to students’ interests and ability levels supports learning by increasing their interest in math and ensuring problems are within their readiness level \cite{baker_results_2020, bernacki_role_2018, nasir_court_2008, pinkard_equitable_2020, walkington_using_2013}. Typically, teachers manually write or curate MWPs customized to students' interests or ability levels. However, with competing responsibilities, limited time, and increasing burnout \cite{kuncl_overworked_2024}, teachers must often rely on standard question sets from educational websites or practice tests for end-of-year examinations, which are limited in quantity, not customized, and not always easily searchable for math standards/topics \cite{vadoe_sol_nodate}. We propose that large language models (LLMs) can address this challenge by automatically generating MWPs aligned with students' interests and ability levels given their vast natural language ability and increasing level of math reasoning \cite{ahn_large_2024}. 

However, given recent work has found LLMs struggle to generate educational MWPs \cite{ariyarathne_elementary_2025,christ_mathwell_2024}, further study must be conducted before directly using them in educational settings. We aim to explore and enhance LLMs' capacity to generate educational MWPs tailored to students' interests and ability levels as indicated by alignment with math standards from the US. Recent work has generated educational MWPs with LLMs customized to students' interests \cite{christ_mathwell_2024}, but not math standards. Other work has generated math practice problems, including MWPs, with LLMs aligned with math topics \cite{ariyarathne_elementary_2025, sun_multi-objective_2025}, but they only generate problems without solutions. Two other works have generated MWPs with LLMs using reference problems \cite{niyarepola_math_2022, zong_solving_2023}, but they too generate problems without solutions. To be useful to students and teachers, generated problems must have solutions written in clear natural language and be aligned with a math standard (e.g., single-step whole number multiplication not exceeding 100) rather than general math topics like multiplication to allow teachers fine-grained control over output MWPs. We refer to this task as standards-aligned educational MWP generation.

To evaluate LLMs' ability to generate standards-aligned educational MWPs, we use a joint human expert and LLM judge approach to assess over 11,000 MWPs generated by LLMs based on four criteria that make MWPs educational and standards-aligned. We adapt the first three criteria, \textit{Solvability}, \textit{Accuracy}, and \textit{Educational Appropriateness}, from \citet{christ_mathwell_2024} and introduce \textit{Standards Alignment}, a new criterion. The nuance needed to simultaneously identify problems that have solutions readable for teachers/students, are educationally appropriate, and effectively incorporate a math standard motivates the need for involving real teachers in evaluations, as it can take teachers many years to develop this expertise. 

Our study has five phases. In phase one, we use a joint expert-LLM judge approach to annotate a small human-written MWP dataset for grade school math standards given there are no datasets fully annotated for these standards. In phase two, we use the labeled data to prompt a LLM to generate over 3,000 MWPs that teachers annotate for our four criteria to expand the amount of training and prompting data for this task. In phase three, we train a mid-size (12B) LLM using the second phase's annotation data to validate our data for training standards-aligned educational MWP generators. In this stage, we further show the quality of our annotated data by using it to train a small text classifier that filters outputs from both our trained model and an existing 30B LLM without further training. In phase four, we evaluate over 8,000 MWPs from our models as well as several open and closed baselines with an LLM judge aligned with feedback from teachers. In phase five, we conduct the first study of customized LLM-generated MWPs with real grade school students.

Through our evaluations, we find performance gaps in the task of generating standards-aligned educational MWPs between open and closed models and between smaller and larger open models. We validate our data by demonstrating both that training on it is sufficient to nearly close the gap in performance between smaller and larger open LLMs and filtering outputs using our text classifier is enough to help a 30B LLM outperform closed baselines without any training. We also show our data and models' outputs are higher quality and more similar to human-written MWPs than outputs from comparison models on several automatic metrics. Finally, we show students prefer our generated MWPs to human-written MWPs while performing similarly on them. Our key contributions are:
\begin{itemize}
    \item We find open models, especially small ones, struggle to generate standards-aligned educational MWPs relative to closed models.
    \item We annotate over 3,000 MWPs with real teachers, which we filter to create the first training dataset for standards-aligned educational MWP generation, the Standards-Targeted Educational Math dataset (STEM).
    \item We use our data to create two models for standards-aligned educational MWP generation that achieve SOTA performance, which we release alongside all of our annotated data.
    \item We conduct the first study of custom LLM-generated MWPs with grade school students, demonstrating their value in K-12 education.
\end{itemize}
\begin{figure*}
    \centering
    \includegraphics[width=1\linewidth]{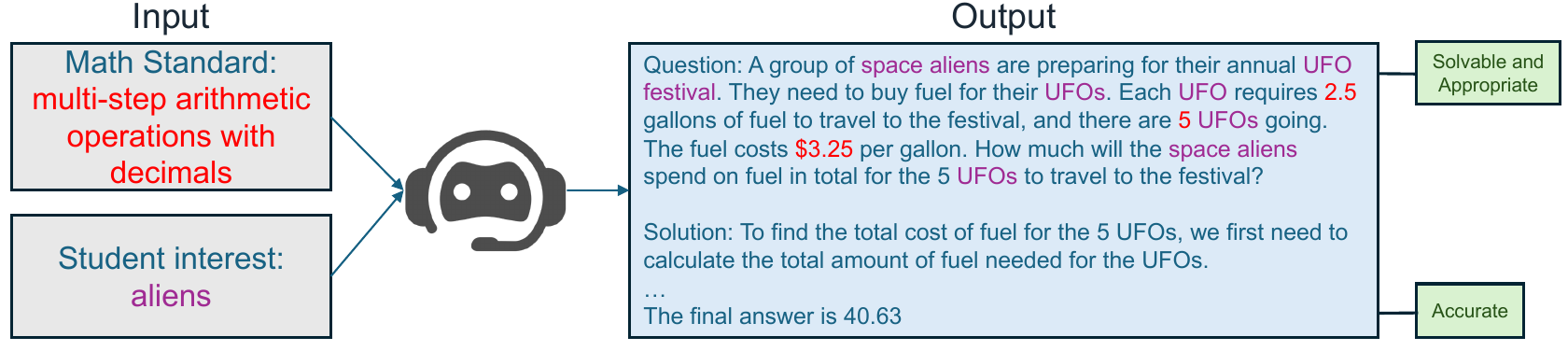}
    \caption{Generating standards-aligned educational MWPs with LLMs. To be educationally useful, problems must be solvable, accurate, educationally appropriate, and aligned with a math standard.}
    \label{fig:EDUMATH_figure1}
\end{figure*}
\section{Related Work}
\paragraph{Math Question/Answer (QA) Datasets} \label{math_qa}
There are several popular datasets for training and evaluating LLMs on grade school math, including GSM8K \cite{cobbe_training_2021}, GSM-Hard \cite{gao_pal_2023}, ASDIV \cite{miao_diverse_2020} and SVAMP \cite{patel_are_2021}. Many works have developed other grade school math benchmark datasets or large training datasets consisting of different solution rationales for existing datasets, particularly GSM8K, including both those that are human written \cite{kim-etal-2023-aint, mishra_lila_2023} and synthetic \cite{li_common_2024,li_synthetic_2024,mitra2024orcamath, pmlr-v202-shi23a, tang_mathscale_2024,toshniwal2024openmath, yu2023metamath, yuan_scaling_2023, yue_mammoth_2023}. 

Because existing datasets are designed to train and evaluate grade school math reasoning, they are not aligned with training standards-aligned educational MWP generators. Training data for such generators must contain high-quality grammar, have both questions and answers, contain readable solutions for teachers and students, be segmented by grade level and standards, have questions that are appropriate for the given grade levels and standards, have questions similar to teacher-written questions, and contain content appropriate for a school setting. We consider high-quality grammar and similarity to MWPs students see in a classroom setting as minimally sufficient criteria any potentially useful training dataset must possess. Existing datasets that contain these minimum sufficiency criteria and are aligned with one or more of the other criteria defined above are GSM-Hard \cite{gao_pal_2023}, GSM8K \cite{cobbe_training_2021}, MathInstruct GSM8K \cite{yue_mammoth_2023}, ASDIV \cite{miao_diverse_2020}, SVAMP \cite{patel_are_2021}, EGSM \cite{christ_mathwell_2024}, and Mathwizards \cite{ariyarathne_elementary_2025}, although none contain all criteria. 
\paragraph{Math Word Problem (MWP) Generation}\label{word_problem_generation}
Closest to our work is \citet{ariyarathne_elementary_2025}, who use LLMs to generate elementary MWPs aligned with a specific math topic like single-digit addition. However, they do not generate solutions, nor do they align their problems with the full language of math standards, which specify both topics and complexity constraints MWPs must follow (e.g., single-step single-digit addition). Further, they do not conduct evaluations involving real teachers and, instead, rely on automated evaluations and evaluations with non-education college students. Another closely related work is \citet{sun_multi-objective_2025}, who use LLMs to generate math problems aligned with a math topic and difficulty level. However, they do not generate solutions, nor are their problems aligned with math standards; instead, they are aligned with general math topics like the Pythagorean theorem. Further, they do not involve teachers in their evaluations and instead rely only on automatic evaluation methods. 

Most other MWP generation works use LLMs or traditional NLP approaches to generate MWPs based on pre-specified equations, re-write existing MWPs, or provide new MWPs based on reference problems \cite{jiao_automatic_2023, koncel-kedziorski_theme-rewriting_2016, kumar_validity_2025, norberg2023rewriting, niyarepola_math_2022,qin_mathematical_2023, qin_math_2024,wang_math_2021, wu_automatic_2022,zhou_towards_2019, zong_solving_2023,zhou2023learning}, each of which limit the originality and range of outputs and can often lead to rephrasing training data. These approaches also only generate problems without solutions and require additional input from users, which is infeasible for teachers or students who want to generate customized MWPs, making them incomparable to our work. To address these issues, we generate MWP question/answer pairs based on a grade level and standard simultaneously without requiring input reference problems or equations. 

\section{Methods}\label{methods}
We define math word problems (MWPs) as natural language questions describing contextual math scenarios paired with solutions written in a natural language, chain-of-thought (CoT) \cite{wei_chain--thought_2023} format. We study generating MWPs with LLMs aligned with a math standard provided in the prompt, as shown in Figure \ref{fig:EDUMATH_figure1}. We evaluate generated MWPs based on the grade levels, standards, evaluation criteria, and automatic metrics defined below and conduct a multi-stage pipeline to generate training data and train models for the task of standards-aligned educational MWP generation (see Appendix \ref{model_pipeline} for a visual of this pipeline). 

\paragraph{Grade Levels and Standards Studied}
We study generating MWPs for grades 3-5 in the US. We choose to study grades 3-5 for two reasons: 1) choosing a narrow grade range allows us to fully assess models' abilities to generate MWPs for all relevant math standards in these grade levels, as each grade has many standards that can be assessed with MWPs, and 2) these are the grade levels for which MWPs are most used and, therefore, where a MWP generator would be most useful \cite{daroczy_word_2015, article, schwartz_why_2023, verschaffel_word_2020}. Expanding to other grade levels is a promising direction for future work but would dilute teacher annotation efforts given a fixed annotation budget. We use MWP math standards from the Virginia Standards of Learning (VA SOL) because they reference bounded difficulty levels and number ranges for all standards, while the Common Core standards, which are used in many US states, are not always as specific \cite{common_core_state_standards_initiative_common_nodate, vdoe_k-12_nodate}. For all grades, VA SOLs are hierarchical and build on one another such that standards encountered later rely on those learned previously \cite{vdoe_k-12_nodate}. We provide a mapping of VA SOLs to Common Core standards in this paper's repo, enabling future work to switch between these standard sets without needing to retrain our models. 

\paragraph{Evaluation Criteria}
We evaluate MWPs on four criteria. The first three criteria are adapted from \citet{christ_mathwell_2024}. \textit{Solvability} evaluates if questions are  possible to solve and have one correct answer. \textit{Accuracy} assesses if generated CoT solutions are accurate. We label MWPs with accurate final answers but inaccurate, confusing, or overly complicated reasoning as inaccurate because our solutions must be readable for teachers and students. \textit{Educational Appropriateness} evaluates if teachers be comfortable giving the MWP to one of their students. An educationally appropriate MWP is one that makes sense, does not contain grammar errors or conflicting information, and is about topics appropriate for 3rd-5th grade students in a school setting. The final criterion we introduce in this work is \textit{Standards Alignment}, or if the MWP effectively incorporates a pre-specified math standard. While the latter three criteria are hard to precisely define, they are similar to those used in other MWP generator works \cite{christ_mathwell_2024, jiao_automatic_2023, qin_mathematical_2023, qin_math_2024} and are meant to model how real teachers evaluate MWPs, which is why emphasize teacher involvement in evaluating MWPs. See Appendix \ref{EDUMATH:sample_annotated_data} for MWPs that do and do not meet these criteria.

\paragraph{Automatic Evaluation Metrics}
Readability automatically assesses the appropriateness of MWPs. We use a composite of 8 readability metrics to comprehensively assess readability: the Automated Readability Index (ARI) \cite{kincaid_derivation_1975}, Flesch-Kincaid Grade Level (FKGL) \cite{flesch_new_1948}, Flesch Reading Ease (FRE) \cite{kincaid_derivation_1975}, CAREC and CAREC-M \cite{crossley_moving_2019}, Coh-Metrix L2 Reading Index (CML2RI) \cite{crossley_assessing_2008}, New Dale-Chall \cite{chall_readability_1995}, and the Simple Measure of Gobbledygook (SMOG) \cite{mc_laughlin_smog_1969}. We use min-max scaling to normalize each metric in the range of [0, 1] and flip the CML2RI and FRE scores so that higher scores represent more difficult to read texts to match the other metrics. We then report the average of these scaled metrics to capture the overall reading complexity of each MWP. In general, lower reading levels are preferred for MWPs because high reading levels can harm student performance, especially for those struggling in math \cite{walkington_how_2018}. We also calculate each MWP's average token length, which is a proxy for complexity given that longer questions often include more mathematical operations. 

Like prior work \cite{christ_mathwell_2024, jiao_automatic_2023,zhou2023learning}, we use Perplexity (PPL) and BERTScore to evaluate the quality of our synthetic MWPs and compare them to human-written MWPs, respectively. Lower PPL suggests higher-quality MWPs, which we calculate with GPT-2. We use BERTScore \cite{zhang2020bertscore} to compute the semantic similarity of synthetic MWPs to each other and compare it to that of human-written MWPs. A higher BERTScore for synthetic MWPs suggests they are more similar to each other than human-written MWPs, and a lower BERTScore suggests the opposite. We compute BERTScore for synthetic MWPs relative to human-written MWPs to identify if they are similar and use the BERTScore of human-written MWPs relative to each other as a reference. For all comparisons to human-written MWPs, we use ASDIV's 3rd-5th grade subset since it is the only human-written English MWP dataset labeled for grade level.

\subsection{Labeling Existing Data for Math Standards}
MWPs with labels for math standards are required for training or prompting models to generate standards-aligned educational MWPs. However, no grade school math question/answer datasets are labeled for standards. ASDIV \cite{miao_diverse_2020} has labels for grade and math topics like addition, but not standards, so we used a four-stage approach to label its 3rd-5th grade MWPs (n = 1,027) for VA SOLs. In stage one, an undergraduate education student labeled MWPs for all relevant VA SOLs, as MWPs may address more than one standard. In stage two, a research team member with K-12 teaching experience checked the labels and made adjustments, often reducing the number of VA SOLs associated with MWPs in cases where the student over-labeled MWPs. We also identified and removed two MWPs that were unsolvable and rewrote several others to make them solvable.

In stage three, we prompted Llama 3.3 70B IT \cite{meta_llama_nodate} with each ASDIV MWP and its VA SOLs from stage two and asked if the question included each VA SOL. See Appendix \ref{EDUMATH:annotate_standards_prompt} for the prompt. We also used a few-shot prompt (see Appendix \ref{EDUMATH:solution_generation_prompt}) for generating readable CoT solutions for the ASDIV subset given it does not have solutions in this format. We compared generated solutions against ground-truth answers to ensure solution accuracy. In stage four, our research team member with K-12 teaching experience reviewed the Llama 3.3 standard labels and made adjustments to remove VA SOLs that were not incorporated in MWPs. To ensure solution quality, we had Gemma 3 27B IT \cite{team_gemma_2025} review generated solutions and rewrite any it deemed were too complex or contained incorrect/unnecessary intermediate reasoning (see Appendix \ref{EDUMATH:solution_revision_prompt} for the prompt). Finally, the undergraduate education student reviewed solutions to ensure intermediate reasoning steps were accurate and the solution was readable for teachers/students. This resulted in a final dataset with 1,025 MWPs with gold labels for VA SOLs and CoT solutions (see Appendix \ref{EDUMATH:asdiv_example} for an example).

\subsection{Synthetic Data Generation and Expert Annotation} \label{STEM_annotation}
While our labeled dataset provides MWPs for prompting, it is not sufficient for training a standards-aligned educational MWP generator given its small size. As shown in Appendix \ref{standards_counts}, most of the 23 VA SOL combinations in the labeled dataset only have a few MWPs they are associated with. Further review of VA SOLs revealed another 15 combinations of standards that can be addressed by MWPs but are not in the labeled dataset. To address these gaps, we generated a large synthetic dataset (n = 3,012) and evaluated it with teachers. We generated these data with Llama 3.3 70B IT \cite{meta_llama_nodate} using the prompt in Appendix \ref{EDUMATH:mwp_generation_prompt}. We generated roughly 90 samples per VA SOL combination that is not in our labeled dataset and roughly 80 samples per VA SOL combination that is in the labeled dataset but has less than 100 samples to ensure coverage of all relevant VA SOLs while emphasizing those that are not already included. \newline \indent Next, teachers annotated generated MWPs through Prolific. Annotators (n = 1,372) identified as US teachers and were paid \$12/hour. Annotators labeled MWPs for solvability, accuracy, educational appropriateness, and standards alignment using the directions in Appendix \ref{EDUMATH:annotator_directions}. See Appendix \ref{EDUMATH:annotator_directions} for demographics and time spent per MWP. Each MWP was annotated twice; if the annotators disagreed on any criteria, the MWP was annotated a third time. We assigned final labels based on majority vote. We labeled a question as meeting all criteria (MaC) if it was labeled as solvable, accurate, educationally appropriate, and standards-aligned. \newline \indent The first two annotators agreed on solvability $90.1 \pm 1.2\%$ of the time, accuracy $76.6 \pm 1.5\%$ of the time, educational appropriateness $77.5 \pm1.5\%$ of the time, standards alignment $75.3 \pm 1.6\%$ of time, and MaC $65.5 \pm 1.7\%$ of the time. The agreement rate for solvability and appropriateness is higher than that reported in \citet{christ_mathwell_2024}, while the agreement rate for accuracy and MaC is lower. The lower accuracy agreement rate could be because accuracy in this work evaluates both final answer and intermediate reasoning accuracy as well as solution readability, whereas \citet{christ_mathwell_2024} only evaluated final answer accuracy. The lower MaC agreement rate could be because it is based on agreement across four metrics instead of three like \citet{christ_mathwell_2024}. However, to ensure only the highest quality data are included in our training dataset, we had each MWP annotated by Gemma 3 27B IT using few-shot examples and the same directions given to annotators (see Appendix \ref{EDUMATH:automatic_annotation_prompt} for the prompt). We flipped the label for MWPs teachers marked as MaC but Gemma 3 27B IT did not, ensuring our final dataset contains only the highest-quality MWPs and reducing potential human labeling errors. We retained the label for any MWP humans identified as not MaC and Gemma 3 27B IT identified as MaC to rely on teachers’ expertise in cases where MWPs do not MaC due to a nuance only human experts can identify. 
\begin{table*}[t]
\centering
\resizebox{\linewidth}{!}{
\begin{tabular}{lcccccccc}
\toprule
\textbf{Dataset}& \textbf{N} &\textbf{\shortstack{Contains\\Answers}}& \textbf{\shortstack{Teacher\\ Annotated}}&\textbf{\shortstack{Readable\\Solutions}}&\textbf{\shortstack{Full Math Standard\\Annotated}}& \textbf{Length}&\textbf{\shortstack{Readability}}&\textbf{BF1}\\ 
\midrule
 Mathwizards& 3,999&\xmark&  \xmark&  \xmark& \xmark& 39.5 (14.5)&0.365 (0.179)&73.0\\
 GSM-Hard& 1,319 &$\checkmark$  & \xmark& \xmark& \xmark& 72.9 (25.6)& 0.367 (0.178)&73.1\\
 GSM8K& 8,792 &$\checkmark$  & \xmark& \xmark& \xmark& 67.0 (24.4)& 0.371 (0.177)&73.1\\
MathInstruct GSM8K& 6,403 &$\checkmark$  &   \xmark&  \xmark&  \xmark& 66.2 (23.9)&0.371 (0.178)&73.2\\
ASDIV& 2,305 &$\checkmark$  &  \xmark&  \xmark&  \xmark& 45.1 (15.8)&0.363 (0.179)&73.6\\
 SVAMP& 1,000 &$\checkmark$  & \xmark& \xmark& \xmark& 47.3 (11.7)& 0.356 (0.178)&75.9\\
 EGSM& 2,093 &$\checkmark$  & $\checkmark$  & \xmark& \xmark& 57.2 (15.7)& 0.358 (0.164)&73.1\\
 STEM (Ours)& 2,577 &$\checkmark$  & $\checkmark$  & $\checkmark$& $\checkmark$  & 53.9 (18.4)& 0.384 (0.177)&73.6\\\bottomrule
\end{tabular}
}
\caption{Characteristics of datasets that can be used to train grade school MWP generators. N is the deduplicated number of questions, Length is average question length (in tokens), readability is measured by a composite of 8 readability metrics, and BF1 is BERTScore F1. Standard deviations, where applicable, are in parentheses.}
\label{tab:EDUMATH_dataset_comparison}
\end{table*}
\subsection{Standards-Targeted Educational Math Dataset}
Our annotation process identified 1,552 MWPs that MaC, and we compile these MWPs along with the labeled ASDIV subset into a new dataset we call Standards-Targeted Educational Math (STEM). Table \ref{tab:EDUMATH_dataset_comparison} shows STEM's advantages over other grade school MWP datasets. Notably, STEM is the only dataset annotated by teachers for math standards and with solutions designed to be readable for students and teachers, making it best suited for training a standards-aligned educational MWP generator. On other metrics, STEM is similar to existing datasets. STEM's question length is in the middle of the range for grade school math datasets, while its average readability is slightly higher than these datasets. STEM's BERTScore is similar to that of human-written datasets, suggesting its MWPs are similar. Overall, these metrics highlight that STEM is similar to human-written data but tailored to standards-aligned educational MWP generation.
\subsection{Finetuning on Annotated Data}
To validate our dataset's quality, we use it to finetune a mid-size open LLM, Gemma 3 12B IT \cite{team_gemma_2025}. We first conduct Supervised Finetuning (SFT) using STEM. Next, we conduct Kahneman-Tversky Optimization (KTO) \cite{ethayarajh_kto_2024} on our SFT model with our annotated data given MaC labels are binary. Next, we train a ModernBERT text classifier \cite{warner_smarter_2024} using our annotation data to label MWPs for MaC that reaches 79\% accuracy and a 0.861 AUC-ROC on a held-out test set. We then use this classifier on top of our trained KTO model to filter MWPs to only those labeled as high quality, resulting in EDUMATH 12B.\footnote{Using a post-processing classifier is akin to how proprietary models reach SOTA performance by using classifiers or multiple models (e.g., \citet{openai_gpt-5_2025})} We also stack our classifier on top of Qwen 3 30B \cite{yang_qwen3_2025} to create EDUMATH 30B. Appendix \ref{finetuning_ablation} shows ablations for EDUMATH 12B's training. Both finetuning stages and automatic filtering using our text classifier significantly improve model performance, showing the quality of our data. See Appendix \ref{EDUMATH_training_hyperparameters} for further training details and hyperparameters.

\section{Evaluating EDUMATH Models} \label{EDUMATH_experiments}
Next, we rigorously evaluate both EDUMATH models using our evaluation criteria and automatic evaluations. We compare 1,000 MWPs from each EDUMATH model to those of open and closed models to see if small models trained on our high-quality data outperform larger, general purpose models. For closed models, we evaluate 250 MWPs from each of GPT-4o \cite{openai_gpt-4o_2024},  GPT-4.1 \cite{openai_introducing_2025}, and GPT-4.5 \cite{openai_introducing_2025}. We use a smaller set of MWPs for closed models due to API cost. For open models, we evaluate 1,000 MWPs from Gemma 3 12B IT \cite{team_gemma_2025}, Gemma 3 27B IT \cite{gemma_team_gemma_2024}, Qwen 3 30B IT \cite{yang_qwen3_2025}, and Qwen 3 235B IT \cite{yang_qwen3_2025}. We prompt each model using 8-shot examples from STEM for a given math standard and ask it to create a new MWP based on the same math standard (see Appendix \ref{EDUMATH:mwp_generation_prompt} for prompting details). Like other MWP generator works \cite{ariyarathne_elementary_2025,christ_mathwell_2024, sun_multi-objective_2025}, this prompting approach ensures fair comparisons across models by providing high-quality examples of MWPs in the expected format and keeping the prompt fixed across models.
\begin{table*}[t]
\centering
\resizebox{\linewidth}{!}{
\begin{tabular}{clcccccccc}
\toprule
& \textbf{Model} & \textbf{PPL $\downarrow$} & \textbf{BF1}& \textbf{ASDIV BF1}& \textbf{\shortstack{Question\\Length}}&\textbf{\shortstack{Solution\\Length}}&\textbf{\shortstack{Question\\Readability}}&\textbf{\shortstack{Solution\\Readability}}&\textbf{MaC} \\
\midrule
 \parbox[t]{2mm}{\multirow{3}{*}{\rotatebox[origin=c]{90}{API}}}&
 GPT-4o& 16.1 (0.33)& 75.2& 74.2&   63.5 (1.10)&152.3 (5.13)& 0.378 (0.004)&0.399 (0.004)&  92.8 (1.6)\\
 & GPT-4.1& 16.3 (0.35)& 75.3& 74.3&   64.5 (1.30)&\textbf{150.2 (4.34)}& 0.385 (0.004)&0.400 (0.004)&92.8 (1.6)\\
 & GPT-4.5& 15.7 (0.33)& 75.0& 74.1&   61.8 (1.09)&150.6 (4.06)& 0.383 (0.004)&0.399 (0.004)&  92.0 (1.7)\\
\midrule
\parbox[t]{2mm}{\multirow{6}{*}{\rotatebox[origin=c]{90}{Public}}}&
Gemma 3 12B IT& 11.3 (0.10)& 77.2&  74.0&   84.7 (0.72)&240.8 (3.49)& 0.372 (0.002)&0.376 (0.002)& 63.9 (1.5)\\
& Gemma 3 27B IT& 12.2 (0.12)& 77.1&  74.5&   76.1 (0.73)&215.1 (3.17)& 0.371 (0.002)&0.378 (0.002)& 75.4 (1.3)\\
 & Qwen 3 30B IT& 12.3 (0.10)& 76.9& 74.0&   68.1 (0.70)&202.9 (3.12)& 0.383 (0.002)&0.383 (0.002)& 87.3 (1.0)\\
 & Qwen 3 235B IT& 12.5 (0.12)& 76.1& 74.1&   66.5 (0.64)&186.5 (2.53)& 0.392 (0.002)&0.378 (0.002)& 89.0 (1.0)\\
 & EDUMATH 12B (Ours)& \textbf{9.5 (0.10)}& \textbf{74.5}& \textbf{73.8}&   \textbf{54.9 (0.54)}&166.7 (2.45)& 0.389 (0.002)&0.381 (0.002)& 85.9 (1.2)\\
& EDUMATH 30B (Ours)& 12.0 (0.13)& 76.1&  \textbf{73.8}&   60.4 (0.71)&163.5 (2.28)& 0.380 (0.002)&0.379 (0.003)& \textbf{94.6 (0.9)*}\\
\bottomrule
\end{tabular}}
\caption{Comparing LLMs for standards-aligned educational MWP generation. PPL is perplexity, BF1 is BERTScore F1, ASDIV BF1 compares each model's MWPs to ASDIV's 3rd-5th grade subset, length is average token length, readability is a composite of 8 metrics, and MaC is meets all criteria. Bold indicates the best performance in each metric (where applicable) and a * indicates the difference between the best open MaC performance and second best open MaC performance is significant at the p < 0.01 level. Standard errors, where applicable, are in parentheses.}
\label{tab:EDUMATH_main_results}
\end{table*}
\subsection{Main Results}
\paragraph{Performance on Evaluation Criteria}
Using our Gemma 3 27B IT annotator, we evaluated each model's MWPs for solvability, accuracy, educational appropriateness, and standards alignment. To reduce LLM calls, we annotate for all criteria together and have the LLM output if the MWP MaC along with its reasoning (see Appendix \ref{EDUMATH:automatic_annotation_prompt} for an example). To validate our labels, our research team member with K-12 teaching experience and undergraduate education student labeled 100 randomly selected samples from our evaluation set. The annotators agreed with each other 76\% of the time on MaC and had an average agreement rate of 75\% with Gemma 3 27B IT for MaC. The Cohen’s  $\kappa $ for MaC between the annotators (0.34) was nearly equivalent to the $\kappa$ for MaC between the annotators and Gemma 3 27B IT (0.30). Both Cohen's Kappas suggest fair agreement, which is expected given MaC is a composite metric indicating agreement across four evaluation criteria and has unbalanced labels. Given the agreement rates and Cohen's Kappas are nearly identical between experts and our automated annotator, we are confident in our labels. 

Table \ref{tab:EDUMATH_main_results} shows the MaC rate for each model. Our 30B model outperforms every baseline in MaC, while our 12B model outperforms Gemma 3 27B IT and nearly matches Qwen 3 30B IT, two larger and more capable models. The table also shows performance gaps between open and closed models and between smaller open and larger open models that our data/models eliminate or nearly eliminate, suggesting our data are sufficient for enabling open models to meet SOTA performance.
\begin{table*}[t]
\centering
\resizebox{\linewidth}{!}{
\small
\begin{tabular}{clcccccccc}
\toprule
& \textbf{Model}& \textbf{Add/Sub}&\textbf{Mult/Div} & \textbf{Fractions}&\textbf{Decimals}&\textbf{\shortstack{Perimeter/\\Area/\\Volume}}&\textbf{Patterns} & \textbf{Time}&\textbf{Conversion}\\ \midrule
\parbox[t]{2mm}{\multirow{3}{*}{\rotatebox[origin=c]{90}{API}}}&
GPT-4o& 91.7&  91.3& 88.9&\textbf{96.2}& 96.4& \textbf{100.0}& 92.3&\textbf{100.0}\\
 & GPT-4.1& 89.4& 91.3& 88.9& \textbf{96.2}& 98.2& 95.0& 92.3&\textbf{100.0}\\
& GPT-4.5& 90.2&  89.3& 82.2&92.3& 94.6& \textbf{100.0}& \textbf{100.0}&\textbf{100.0}\\
\midrule
\parbox[t]{2mm}{\multirow{6}{*}{\rotatebox[origin=c]{90}{Public}}}&
Gemma 3 12B IT& 63.0& 59.2& 61.5&62.2&79.4& 77.0& 67.3&66.7\\
& Gemma 3 27B IT& 71.5& 69.9& 82.3&77.7&80.5& 90.2& 85.2&74.5\\
 & Qwen 3 30B IT& 82.1&  81.9& 91.1&87.0& 93.6& 97.6& 93.5&88.7\\
 & Qwen 3 235B IT& 84.0&  83.6& 89.5&91.7& 97.3& 97.6& 94.4&94.4\\
 & EDUMATH 12B (Ours)& 88.0& 81.0& 78.6&93.0&92.9& 84.8& 98.1&92.5\\
& EDUMATH 30B (Ours)& \textbf{94.3}& \textbf{92.0}& \textbf{92.2}&95.2&\textbf{98.8}& 97.5& 93.5&\textbf{100.0}\\
\bottomrule
\end{tabular}
}
\caption{Meets all criteria rate for each model by topic. Bold indicates the best performance for each topic.}
\label{tab:errors_by_topic_EDUMATH}
\end{table*}
\paragraph{Comparison to Human-written MWPs}
Table \ref{tab:EDUMATH_main_results} also shows PPL, BERTScore, and the average question and solution length and readability for each model's MWPs. Our 12B model has the lowest PPL among all models and our 30B model has the lowest PPL among larger models, suggesting their MWPs are high quality. Our 12B model's BERTScore is most similar to ASDIV's within-dataset BERTScore among all models, while our 30B model ties with Qwen 3 235B IT in having a BERTScore closest to ASDIV's among the open models excluding EDUMATH 12B. The BERTScores of our models' MWPs compared to ASDIV are roughly equivalent to ASDIV's within-dataset BERTScore, suggesting their MWPs are similar in diversity to human-written MWPs, while the MWPs from other models have higher BERTScores for this comparison. Our models' average question lengths are most similar to those of STEM and ASDIV, suggesting their complexity, as measured by this proxy, is similar to teacher-written MWPs. Our models have the shortest solutions among open models, though the closed baselines have the shortest solutions among all models. Shorter solutions are preferable to longer, more verbose ones that may be daunting to read.

For readability, all models output questions with readability similar to STEM. For solution readability, all open models' solutions have lower readability than STEM, while closed models exceed STEM's readability, denoting a strength of open models. Overall, Table \ref{tab:EDUMATH_main_results} suggests the EDUMATH models' MWPs are high quality and similar to human-written MWPs. The alignment between our models' performance across the automated evaluation criteria and higher rates of MaC further indicates our human evaluation criteria effectively identify high-quality MWPs.

\paragraph{Performance by Math Topic}
Although there are 38 math standard combinations, they can be grouped into combinations of one or more of 8 math topics: addition/subtraction, multiplication/division, fractions, decimals, perimeter/area/volume, patterns, elapsed time, and measurement conversion. Exploring MaC rates by topic can highlight the topics each model struggles with and succeeds at. Table \ref{tab:errors_by_topic_EDUMATH} shows this comparison, while Appendix \ref{error_analysis} shows further analysis of common error types for each model. EDUMATH 12B performs best at time problems and struggles the most with fraction problems, while EDUMATH 30B performs best at conversion problems and struggles the most with multiplication/division problems. EDUMATH 30B is the only model for which performance on each topic is above 90\%, further highlighting its SOTA performance. 

\section{Evaluating MWPs with Students} \label{student_study}
While we show our models' MWPs are high quality and can be customized to student interests (see Appendix \ref{customizing_mwps}), it is critical to ensure students perform similarly on their MWPs relative to human-written MWPs and to identify if students prefer their customized MWPs to non-customized MWPs. To test this, 3rd-5th graders solved a worksheet with one MWP generated by EDUMATH 30B and aligned with topics they are interested in and one written by humans, either from existing data (\citet{noauthor_free_nodate} or ASDIV's 3rd-5th grade subset) or written by our research team member with K-12 teaching experience if a worksheet's math topic was not captured in existing data (see Appendix \ref{sample_worksheet} for an example). We also asked students which MWP they preferred and why. We tested these worksheets in two experimental settings at two elementary schools (see Appendix \ref{student_demographics} for school demographics). All worksheets were screened by teachers to ensure the MWPs were appropriate. At the first school, 4th graders (n = 82) solved one worksheet for each of four weeks where both the LLM-generated and human-written MWP were the same for all students. Each MWP was aligned with a standard students were currently learning and the LLM-generated MWP was based on a topic teachers identified as interesting to students. At the second school, 3rd-5th graders (n = 12) receiving services through a math interventionist completed one or two worksheets with the same human-written MWP for every student in each grade level and a customized LLM-generated MWP for each student in that grade level, depending on how many standards they were learning (see Appendix \ref{student_demographics} for details). To inform the customized MWPs, the interventionist surveyed students to identify topics they were interested in, resulting in deep customization for each student. These two experimental conditions allow us to test differential effects of customization at the class and individual level, while also testing our MWPs across all grade levels we study.

In this experiment (see Appendix \ref{student_results} for full results), students performed similarly on human-written and LLM-generated MWPs, further suggesting EDUMATH 30B's MWPs are human quality. Students preferred the LLM-generated MWPs over the human-written MWPs in both conditions. In School 2 (customization at the individual level), every student but one preferred the LLM-generated MWPs. Students most often reported liking the LLM-generated MWP because they liked the topic it was about. These results offer compelling evidence that LLMs can support math education by providing customized MWPs that excite students.

\section{Conclusion}
We explore standards-aligned educational MWP generation and create the first dataset to train models for this task and the only one verified by teachers to have readable solutions. We also evaluate existing models at this task and fully release our annotations to aid future research. We demonstrate the quality of STEM and our annotated data by using them to develop the EDUMATH models, the first standards-aligned educational MWP generators. Our evaluations show EDUMATH 12B matches the performance of larger and more capable open models, while EDUMATH 30B sets a new SOTA. We also conduct the first study of custom LLM-generated MWPs with grade school students, finding they perform similarly on our MWPs relative to human-written MWPS, while preferring our MWPs. Our findings suggest LLMs can contribute to K-12 education by generating customized practice problems to reduce teacher burdens. Future research should develop models and data for other grade levels and types of math problems.

\section*{Limitations}
One limitation of this work is that we restricted it to the study of 3rd-5th grade MWPs. While these are the grade levels for which MWPs are most highly used, all grade levels use MWPs to some degree, and we believe other grade levels are therefore compelling areas for future work. Another limitation of our work is that our MWPs are text-only, and many MWPs students encounter are multi-modal, containing both images/tables/figures and text. Generating these multi-modal MWPs is an interesting and challenging direction for future work. While we use a standard prompt for all evaluated models to make fair comparisons across models, future work could examine prompt engineering approaches to improve the performance of specific models. The high cost of human annotation is a limitation of MWP generator studies broadly, and we hope our annotations, trained text classifier, and Gemma 3 27B IT automated annotator can help motivate future work in automatic classification efforts. Lastly, while we aligned our Gemma 3 27B IT automated annotator with human experts over multiple rounds of iteration and reached an average agreement rate on par with that of the agreement rate between human experts, it is possible the model missed nuances in MWPs that only teachers with years of experience can identify. Future work should thus continue to involve teachers in the evaluation process of model outputs.

\section*{Ethics Statement}
All data used in this study come from open-access datasets and, therefore, should not contain any private sensitive information. EDUMATH may generate questions that are not educationally appropriate and further research should be conducted before deploying the model directly in classroom contexts. 
\bibliography{custom}

@misc{li_common_2024,
	title = {Common {7B} {Language} {Models} {Already} {Possess} {Strong} {Math} {Capabilities}},
	url = {http://arxiv.org/abs/2403.04706},
	urldate = {2024-03-21},
	publisher = {arXiv},
	author = {Li, Chen and Wang, Weiqi and Hu, Jingcheng and Wei, Yixuan and Zheng, Nanning and Hu, Han and Zhang, Zheng and Peng, Houwen},
	month = mar,
	year = {2024},
	note = {arXiv:2403.04706 [cs]},
}

@inproceedings{koncel-kedziorski_theme-rewriting_2016,
	address = {Austin, Texas},
	title = {A {Theme}-{Rewriting} {Approach} for {Generating} {Algebra} {Word} {Problems}},
	url = {https://aclanthology.org/D16-1168},
	doi = {10.18653/v1/D16-1168},
	urldate = {2024-03-29},
	booktitle = {Proceedings of the 2016 {Conference} on {Empirical} {Methods} in {Natural} {Language} {Processing}},
	publisher = {Association for Computational Linguistics},
	author = {Koncel-Kedziorski, Rik and Konstas, Ioannis and Zettlemoyer, Luke and Hajishirzi, Hannaneh},
	editor = {Su, Jian and Duh, Kevin and Carreras, Xavier},
	month = nov,
	year = {2016},
	pages = {1617--1628},
}

@article{sun_multi-objective_2025,
	title = {Multi-objective math problem generation using large language model through an adaptive multi-level retrieval augmentation framework},
	volume = {119},
	issn = {1566-2535},
	url = {https://www.sciencedirect.com/science/article/pii/S1566253525001101},
	doi = {10.1016/j.inffus.2025.103037},
	urldate = {2025-03-03},
	journal = {Information Fusion},
	author = {Sun, Jianwen and Shi, Wangzi and Shen, Xiaoxuan and Liu, Shengyingjie and Wei, Luona and Wan, Qian},
	month = jul,
	year = {2025},
	pages = {103037},
}

@article{kumar_validity_2025,
	title = {Validity {Checking} and {Repairing} of {Machine} {Generated} {Transfer}-{Type} {Word} {Problems}},
	issn = {1570-5838},
	url = {https://journals.sagepub.com/doi/abs/10.1177/15705838241303829},
	doi = {10.1177/15705838241303829},
	language = {en},
	urldate = {2025-02-03},
	journal = {Applied Ontology},
	author = {Kumar, Suresh and Sreenivasa Kumar, P},
	month = jan,
	year = {2025},
	note = {Publisher: SAGE Publications},
	pages = {15705838241303829},
}

@misc{gemma_team_gemma_2024,
	title = {Gemma 2: {Improving} {Open} {Language} {Models} at a {Practical} {Size}},
	shorttitle = {Gemma 2},
	url = {http://arxiv.org/abs/2408.00118},
	doi = {10.48550/arXiv.2408.00118},
	urldate = {2024-09-30},
	publisher = {arXiv},
	author = {Gemma Team and Riviere, Morgane and Pathak, Shreya and Sessa, Pier Giuseppe and Hardin, Cassidy and Bhupatiraju, Surya and Hussenot, Léonard and Mesnard, Thomas and Shahriari, Bobak and Ramé, Alexandre and Ferret, Johan and Liu, Peter and Tafti, Pouya and Friesen, Abe and Casbon, Michelle and Ramos, Sabela and Kumar, Ravin and Lan, Charline Le and Jerome, Sammy and Tsitsulin, Anton and Vieillard, Nino and Stanczyk, Piotr and Girgin, Sertan and Momchev, Nikola and Hoffman, Matt and Thakoor, Shantanu and Grill, Jean-Bastien and Neyshabur, Behnam and Bachem, Olivier and Walton, Alanna and Severyn, Aliaksei and Parrish, Alicia and Ahmad, Aliya and Hutchison, Allen and Abdagic, Alvin and Carl, Amanda and Shen, Amy and Brock, Andy and Coenen, Andy and Laforge, Anthony and Paterson, Antonia and Bastian, Ben and Piot, Bilal and Wu, Bo and Royal, Brandon and Chen, Charlie and Kumar, Chintu and Perry, Chris and Welty, Chris and Choquette-Choo, Christopher A. and Sinopalnikov, Danila and Weinberger, David and Vijaykumar, Dimple and Rogozińska, Dominika and Herbison, Dustin and Bandy, Elisa and Wang, Emma and Noland, Eric and Moreira, Erica and Senter, Evan and Eltyshev, Evgenii and Visin, Francesco and Rasskin, Gabriel and Wei, Gary and Cameron, Glenn and Martins, Gus and Hashemi, Hadi and Klimczak-Plucińska, Hanna and Batra, Harleen and Dhand, Harsh and Nardini, Ivan and Mein, Jacinda and Zhou, Jack and Svensson, James and Stanway, Jeff and Chan, Jetha and Zhou, Jin Peng and Carrasqueira, Joana and Iljazi, Joana and Becker, Jocelyn and Fernandez, Joe and van Amersfoort, Joost and Gordon, Josh and Lipschultz, Josh and Newlan, Josh and Ji, Ju-yeong and Mohamed, Kareem and Badola, Kartikeya and Black, Kat and Millican, Katie and McDonell, Keelin and Nguyen, Kelvin and Sodhia, Kiranbir and Greene, Kish and Sjoesund, Lars Lowe and Usui, Lauren and Sifre, Laurent and Heuermann, Lena and Lago, Leticia and McNealus, Lilly and Soares, Livio Baldini and Kilpatrick, Logan and Dixon, Lucas and Martins, Luciano and Reid, Machel and Singh, Manvinder and Iverson, Mark and Görner, Martin and Velloso, Mat and Wirth, Mateo and Davidow, Matt and Miller, Matt and Rahtz, Matthew and Watson, Matthew and Risdal, Meg and Kazemi, Mehran and Moynihan, Michael and Zhang, Ming and Kahng, Minsuk and Park, Minwoo and Rahman, Mofi and Khatwani, Mohit and Dao, Natalie and Bardoliwalla, Nenshad and Devanathan, Nesh and Dumai, Neta and Chauhan, Nilay and Wahltinez, Oscar and Botarda, Pankil and Barnes, Parker and Barham, Paul and Michel, Paul and Jin, Pengchong and Georgiev, Petko and Culliton, Phil and Kuppala, Pradeep and Comanescu, Ramona and Merhej, Ramona and Jana, Reena and Rokni, Reza Ardeshir and Agarwal, Rishabh and Mullins, Ryan and Saadat, Samaneh and Carthy, Sara Mc and Perrin, Sarah and Arnold, Sébastien M. R. and Krause, Sebastian and Dai, Shengyang and Garg, Shruti and Sheth, Shruti and Ronstrom, Sue and Chan, Susan and Jordan, Timothy and Yu, Ting and Eccles, Tom and Hennigan, Tom and Kocisky, Tomas and Doshi, Tulsee and Jain, Vihan and Yadav, Vikas and Meshram, Vilobh and Dharmadhikari, Vishal and Barkley, Warren and Wei, Wei and Ye, Wenming and Han, Woohyun and Kwon, Woosuk and Xu, Xiang and Shen, Zhe and Gong, Zhitao and Wei, Zichuan and Cotruta, Victor and Kirk, Phoebe and Rao, Anand and Giang, Minh and Peran, Ludovic and Warkentin, Tris and Collins, Eli and Barral, Joelle and Ghahramani, Zoubin and Hadsell, Raia and Sculley, D. and Banks, Jeanine and Dragan, Anca and Petrov, Slav and Vinyals, Oriol and Dean, Jeff and Hassabis, Demis and Kavukcuoglu, Koray and Farabet, Clement and Buchatskaya, Elena and Borgeaud, Sebastian and Fiedel, Noah and Joulin, Armand and Kenealy, Kathleen and Dadashi, Robert and Andreev, Alek},
	month = aug,
	year = {2024},
	note = {arXiv:2408.00118 [cs]},
}

@misc{cobbe_training_2021,
	title = {Training {Verifiers} to {Solve} {Math} {Word} {Problems}},
	url = {http://arxiv.org/abs/2110.14168},
	doi = {10.48550/arXiv.2110.14168},
	urldate = {2023-09-09},
	publisher = {arXiv},
	author = {Cobbe, Karl and Kosaraju, Vineet and Bavarian, Mohammad and Chen, Mark and Jun, Heewoo and Kaiser, Lukasz and Plappert, Matthias and Tworek, Jerry and Hilton, Jacob and Nakano, Reiichiro and Hesse, Christopher and Schulman, John},
	month = nov,
	year = {2021},
	note = {arXiv:2110.14168 [cs]},
}

@misc{ahn_large_2024,
	title = {Large {Language} {Models} for {Mathematical} {Reasoning}: {Progresses} and {Challenges}},
	shorttitle = {Large {Language} {Models} for {Mathematical} {Reasoning}},
	url = {http://arxiv.org/abs/2402.00157},
	language = {en},
	urldate = {2024-10-14},
	publisher = {arXiv},
	author = {Ahn, Janice and Verma, Rishu and Lou, Renze and Liu, Di and Zhang, Rui and Yin, Wenpeng},
	month = sep,
	year = {2024},
	note = {arXiv:2402.00157 [cs]},
}

@inproceedings{christ_mathwell_2024,
	address = {Miami, Florida, USA},
	title = {{MATHWELL}: {Generating} {Educational} {Math} {Word} {Problems} {Using} {Teacher} {Annotations}},
	shorttitle = {{MATHWELL}},
	url = {https://aclanthology.org/2024.findings-emnlp.696},
	urldate = {2024-11-10},
	booktitle = {Findings of the {Association} for {Computational} {Linguistics}: {EMNLP} 2024},
	publisher = {Association for Computational Linguistics},
	author = {Christ, Bryan R and Kropko, Jonathan and Hartvigsen, Thomas},
	editor = {Al-Onaizan, Yaser and Bansal, Mohit and Chen, Yun-Nung},
	month = nov,
	year = {2024},
	pages = {11914--11938},
}

@misc{wei_chain--thought_2023,
	title = {Chain-of-{Thought} {Prompting} {Elicits} {Reasoning} in {Large} {Language} {Models}},
	url = {http://arxiv.org/abs/2201.11903},
	doi = {10.48550/arXiv.2201.11903},
	urldate = {2023-03-26},
	publisher = {arXiv},
	author = {Wei, Jason and Wang, Xuezhi and Schuurmans, Dale and Bosma, Maarten and Ichter, Brian and Xia, Fei and Chi, Ed and Le, Quoc and Zhou, Denny},
	month = jan,
	year = {2023},
	note = {arXiv:2201.11903 [cs]},
}

@misc{gao_pal_2023,
	title = {{PAL}: {Program}-aided {Language} {Models}},
	shorttitle = {{PAL}},
	url = {http://arxiv.org/abs/2211.10435},
	doi = {10.48550/arXiv.2211.10435},
	urldate = {2023-07-07},
	publisher = {arXiv},
	author = {Gao, Luyu and Madaan, Aman and Zhou, Shuyan and Alon, Uri and Liu, Pengfei and Yang, Yiming and Callan, Jamie and Neubig, Graham},
	month = jan,
	year = {2023},
	note = {arXiv:2211.10435 [cs]},
}

@misc{yue_mammoth_2023,
	title = {{MAmmoTH}: {Building} {Math} {Generalist} {Models} through {Hybrid} {Instruction} {Tuning}},
	shorttitle = {{MAmmoTH}},
	url = {http://arxiv.org/abs/2309.05653},
	urldate = {2023-12-15},
	publisher = {arXiv},
	author = {Yue, Xiang and Qu, Xingwei and Zhang, Ge and Fu, Yao and Huang, Wenhao and Sun, Huan and Su, Yu and Chen, Wenhu},
	month = oct,
	year = {2023},
	note = {arXiv:2309.05653 [cs]},
}

@inproceedings{niyarepola_math_2022,
	address = {Waterville, Maine, USA and virtual meeting},
	title = {Math {Word} {Problem} {Generation} with {Multilingual} {Language} {Models}},
	url = {https://aclanthology.org/2022.inlg-main.12},
	doi = {10.18653/v1/2022.inlg-main.12},
	urldate = {2023-12-20},
	booktitle = {Proceedings of the 15th {International} {Conference} on {Natural} {Language} {Generation}},
	publisher = {Association for Computational Linguistics},
	author = {Niyarepola, Kashyapa and Athapaththu, Dineth and Ekanayake, Savindu and Ranathunga, Surangika},
	editor = {Shaikh, Samira and Ferreira, Thiago and Stent, Amanda},
	month = jul,
	year = {2022},
	pages = {144--155},
}

@inproceedings{miao_diverse_2020,
	address = {Online},
	title = {A {Diverse} {Corpus} for {Evaluating} and {Developing} {English} {Math} {Word} {Problem} {Solvers}},
	url = {https://aclanthology.org/2020.acl-main.92},
	doi = {10.18653/v1/2020.acl-main.92},
	urldate = {2023-12-20},
	booktitle = {Proceedings of the 58th {Annual} {Meeting} of the {Association} for {Computational} {Linguistics}},
	publisher = {Association for Computational Linguistics},
	author = {Miao, Shen-yun and Liang, Chao-Chun and Su, Keh-Yih},
	editor = {Jurafsky, Dan and Chai, Joyce and Schluter, Natalie and Tetreault, Joel},
	month = jul,
	year = {2020},
	pages = {975--984},
}

@misc{patel_are_2021,
	title = {Are {NLP} {Models} really able to {Solve} {Simple} {Math} {Word} {Problems}?},
	url = {http://arxiv.org/abs/2103.07191},
	doi = {10.48550/arXiv.2103.07191},
	urldate = {2024-01-02},
	publisher = {arXiv},
	author = {Patel, Arkil and Bhattamishra, Satwik and Goyal, Navin},
	month = apr,
	year = {2021},
	note = {arXiv:2103.07191 [cs]},
}

@misc{wang_math_2021,
	title = {Math {Word} {Problem} {Generation} with {Mathematical} {Consistency} and {Problem} {Context} {Constraints}},
	url = {http://arxiv.org/abs/2109.04546},
	doi = {10.48550/arXiv.2109.04546},
	urldate = {2024-01-09},
	publisher = {arXiv},
	author = {Wang, Zichao and Lan, Andrew S. and Baraniuk, Richard G.},
	month = sep,
	year = {2021},
	note = {arXiv:2109.04546 [cs]},
}

@inproceedings{zhou_towards_2019,
	address = {Tokyo, Japan},
	title = {Towards {Generating} {Math} {Word} {Problems} from {Equations} and {Topics}},
	url = {https://aclanthology.org/W19-8661},
	doi = {10.18653/v1/W19-8661},
	urldate = {2024-01-09},
	booktitle = {Proceedings of the 12th {International} {Conference} on {Natural} {Language} {Generation}},
	publisher = {Association for Computational Linguistics},
	author = {Zhou, Qingyu and Huang, Danqing},
	editor = {van Deemter, Kees and Lin, Chenghua and Takamura, Hiroya},
	month = oct,
	year = {2019},
	pages = {494--503},
}

@inproceedings{qin_mathematical_2023,
	address = {New York, NY, USA},
	series = {{SIGIR} '23},
	title = {A {Mathematical} {Word} {Problem} {Generator} with {Structure} {Planning} and {Knowledge} {Enhancement}},
	isbn = {978-1-4503-9408-6},
	url = {https://dl.acm.org/doi/10.1145/3539618.3591937},
	doi = {10.1145/3539618.3591937},
	urldate = {2024-01-09},
	booktitle = {Proceedings of the 46th {International} {ACM} {SIGIR} {Conference} on {Research} and {Development} in {Information} {Retrieval}},
	publisher = {Association for Computing Machinery},
	author = {Qin, Longhu and Liu, Jiayu and Huang, Zhenya and Zhang, Kai and Liu, Qi and Jin, Binbin and Chen, Enhong},
	month = jul,
	year = {2023},
	pages = {1750--1754},
}

@inproceedings{jiao_automatic_2023,
	address = {Cham},
	series = {Lecture {Notes} in {Computer} {Science}},
	title = {Automatic {Educational} {Question} {Generation} with {Difficulty} {Level} {Controls}},
	isbn = {978-3-031-36272-9},
	doi = {10.1007/978-3-031-36272-9_39},
	language = {en},
	booktitle = {Artificial {Intelligence} in {Education}},
	publisher = {Springer Nature Switzerland},
	author = {Jiao, Ying and Shridhar, Kumar and Cui, Peng and Zhou, Wangchunshu and Sachan, Mrinmaya},
	editor = {Wang, Ning and Rebolledo-Mendez, Genaro and Matsuda, Noboru and Santos, Olga C. and Dimitrova, Vania},
	year = {2023},
	pages = {476--488},
}

@article{flesch_new_1948,
	title = {A new readability yardstick},
	volume = {32},
	issn = {0021-9010},
	doi = {10.1037/h0057532},
	language = {eng},
	number = {3},
	journal = {The Journal of Applied Psychology},
	author = {Flesch, R.},
	month = jun,
	year = {1948},
	pmid = {18867058},
	pages = {221--233},
}

@article{walkington_using_2013,
	title = {Using adaptive learning technologies to personalize instruction to student interests: {The} impact of relevant contexts on performance and learning outcomes},
	volume = {105},
	issn = {1939-2176},
	shorttitle = {Using adaptive learning technologies to personalize instruction to student interests},
	doi = {10.1037/a0031882},
	number = {4},
	journal = {Journal of Educational Psychology},
	author = {Walkington, Candace A.},
	year = {2013},
	note = {Place: US
Publisher: American Psychological Association},
	pages = {932--945},
}

@article{bernacki_role_2018,
	title = {The role of situational interest in personalized learning},
	volume = {110},
	issn = {1939-2176},
	doi = {10.1037/edu0000250},
	number = {6},
	journal = {Journal of Educational Psychology},
	author = {Bernacki, Matthew L. and Walkington, Candace},
	year = {2018},
	note = {Place: US
Publisher: American Psychological Association},
	pages = {864--881},
}

@techreport{kincaid_derivation_1975,
	address = {Fort Belvoir, VA},
	title = {Derivation of {New} {Readability} {Formulas} ({Automated} {Readability} {Index}, {Fog} {Count} and {Flesch} {Reading} {Ease} {Formula}) for {Navy} {Enlisted} {Personnel}:},
	shorttitle = {Derivation of {New} {Readability} {Formulas} ({Automated} {Readability} {Index}, {Fog} {Count} and {Flesch} {Reading} {Ease} {Formula}) for {Navy} {Enlisted} {Personnel}},
	url = {http://www.dtic.mil/docs/citations/ADA006655},
	language = {en},
	urldate = {2024-01-15},
	institution = {Defense Technical Information Center},
	author = {Kincaid, J. P. and Fishburne, Jr. and Robert P., Rogers and Richard L., Chissom and {Brad S.}},
	month = feb,
	year = {1975},
	doi = {10.21236/ADA006655},
}

@article{verschaffel_word_2020,
	title = {Word {Problems} in {Mathematics} {Education}: {A} {Survey}},
	volume = {52},
	issn = {1863-9690},
	shorttitle = {Word {Problems} in {Mathematics} {Education}},
	doi = {10.1007/s11858-020-01130-4},
	language = {en},
	number = {1},
	urldate = {2024-01-11},
	journal = {ZDM: The International Journal on Mathematics Education},
	author = {Verschaffel, Lieven and Schukajlow, Stanislaw and Star, Jon and Van Dooren, Wim},
	month = apr,
	year = {2020},
	note = {Publisher: Springer
ERIC Number: EJ1243930},
	pages = {1--16},
}

@article{zong_solving_2023,
	title = {Solving {Math} {Word} {Problems} concerning {Systems} of {Equations} with {GPT}-3},
	volume = {37},
	copyright = {Copyright (c) 2023 Association for the Advancement of Artificial Intelligence},
	issn = {2374-3468},
	url = {https://ojs.aaai.org/index.php/AAAI/article/view/26896},
	doi = {10.1609/aaai.v37i13.26896},
	language = {en},
	number = {13},
	urldate = {2024-01-09},
	journal = {Proceedings of the AAAI Conference on Artificial Intelligence},
	author = {Zong, Mingyu and Krishnamachari, Bhaskar},
	month = sep,
	year = {2023},
	note = {Number: 13},
	pages = {15972--15979},
}

@article{norberg2023rewriting,
  title={Rewriting Math Word Problems with Large Language Models},
  author={Norberg, Kole and Almoubayyed, Husni and Fancsali, Stephen E and De Ley, Logan and Weldon, Kyle and Murphy, April and Ritter, Steve},
  year={2023}, 
  journal = {Proceedings of the Workshop on Empowering Education with LLMs-the Next-Gen Interface and Content Generation 2023 co-located with 24th International Conference on Artificial Intelligence in Education (AIED 2023)}, 
pages = {163--172}, 
address = {Tokyo, Japan},
volume  = {3487}
}

@misc{zhou2023learning,
      title={Learning by Analogy: Diverse Questions Generation in Math Word Problem}, 
      author={Zihao Zhou and Maizhen Ning and Qiufeng Wang and Jie Yao and Wei Wang and Xiaowei Huang and Kaizhu Huang},
      year={2023},
      eprint={2306.09064},
      archivePrefix={arXiv},
      primaryClass={cs.CL}
}

@misc{zhang2020bertscore,
      title={BERTScore: Evaluating Text Generation with BERT}, 
      author={Tianyi Zhang and Varsha Kishore and Felix Wu and Kilian Q. Weinberger and Yoav Artzi},
      year={2020},
      eprint={1904.09675},
      archivePrefix={arXiv},
      primaryClass={cs.CL}
}

@book{chall_readability_1995,
	title = {Readability {Revisited}: {The} {New} {Dale}-{Chall} {Readability} {Formula}},
	isbn = {978-1-57129-008-3},
	shorttitle = {Readability {Revisited}},
	language = {en},
	publisher = {Brookline Books},
	author = {Chall, Jeanne Sternlicht and Dale, Edgar},
	year = {1995},
	note = {Google-Books-ID: 2nbuAAAAMAAJ},
}

@misc{noauthor_free_nodate,
	title = {Free worksheets},
        author= {K5Learning},
	url = {https://www.k5learning.com/},
	language = {en},
	urldate = {2024-02-11},
	journal = {K5 Learning},
}

@misc{vdoe_k-12_nodate,
  author       = {{Virginia Department of Education}},
  title        = {K-12 Standards \& Instruction},
  howpublished = {\url{https://www.doe.virginia.gov/teaching-learning-assessment/instruction}},
  note         = {Accessed: 2025-04-07},
}

@article{article,
author = {Pearce, Daniel and Bruun, Faye and Skinner, Kim and Lopez-Mohler, Claricia},
year = {2013},
month = {04},
pages = {3-19},
title = {What Teachers Say About Student Difficulties Solving Mathematical Word Problems in Grades 2-5},
volume = {8},
journal = {International Electronic Journal of Mathematics Education},
doi = {10.29333/iejme/271}
}

@article{daroczy_word_2015,
	title = {Word problems: a review of linguistic and numerical factors contributing to their difficulty},
	volume = {6},
	issn = {1664-1078},
	shorttitle = {Word problems},
	url = {https://www.ncbi.nlm.nih.gov/pmc/articles/PMC4381502/},
	doi = {10.3389/fpsyg.2015.00348},
	urldate = {2024-02-14},
	journal = {Frontiers in Psychology},
	author = {Daroczy, Gabriella and Wolska, Magdalena and Meurers, Walt Detmar and Nuerk, Hans-Christoph},
	month = apr,
	year = {2015},
	pmid = {25883575},
	pmcid = {PMC4381502},
	pages = {348},
}

@article{schwartz_why_2023,
	chapter = {Teaching \& Learning, Mathematics},
	title = {Why {Word} {Problems} {Are} {Such} a {Struggle} for {Students}—{And} {What} {Teachers} {Can} {Do}},
	issn = {0277-4232},
	url = {https://www.edweek.org/teaching-learning/why-word-problems-are-such-a-struggle-for-students-and-what-teachers-can-do/2023/05},
	language = {en},
	urldate = {2024-02-14},
	journal = {Education Week},
	author = {Schwartz, Sarah},
	month = may,
	year = {2023},
}

@article{qin_math_2024,
	title = {Math {Word} {Problem} {Generation} via {Disentangled} {Memory} {Retrieval}},
	issn = {1556-4681},
	url = {https://dl.acm.org/doi/10.1145/3639569},
	doi = {10.1145/3639569},
	urldate = {2024-02-26},
	journal = {ACM Transactions on Knowledge Discovery from Data},
	author = {Qin, Wei and Wang, Xiaowei and Hu, Zhenzhen and Wang, Lei and Lan, Yunshi and Hong, Richang},
	month = jan,
	year = {2024},
	note = {Just Accepted},
}

@misc{mitra2024orcamath,
      title={Orca-Math: Unlocking the potential of SLMs in Grade School Math}, 
      author={Arindam Mitra and Hamed Khanpour and Corby Rosset and Ahmed Awadallah},
      year={2024},
      eprint={2402.14830},
      archivePrefix={arXiv},
      primaryClass={cs.CL}
}

@inproceedings{kim-etal-2023-aint,
    title = "It Ain{'}t Over: A Multi-aspect Diverse Math Word Problem Dataset",
    author = "Kim, Jiwoo  and
      Kim, Youngbin  and
      Baek, Ilwoong  and
      Bak, JinYeong  and
      Lee, Jongwuk",
    editor = "Bouamor, Houda  and
      Pino, Juan  and
      Bali, Kalika",
    booktitle = "Proceedings of the 2023 Conference on Empirical Methods in Natural Language Processing",
    month = dec,
    year = "2023",
    address = "Singapore",
    publisher = "Association for Computational Linguistics",
    url = "https://aclanthology.org/2023.emnlp-main.927",
    doi = "10.18653/v1/2023.emnlp-main.927",
    pages = "14984--15011",
}

@misc{yu2023metamath,
      title={MetaMath: Bootstrap Your Own Mathematical Questions for Large Language Models}, 
      author={Longhui Yu and Weisen Jiang and Han Shi and Jincheng Yu and Zhengying Liu and Yu Zhang and James T. Kwok and Zhenguo Li and Adrian Weller and Weiyang Liu},
      year={2023},
      eprint={2309.12284},
      archivePrefix={arXiv},
      primaryClass={cs.CL}
}

@InProceedings{pmlr-v202-shi23a,
  title = 	 {Large Language Models Can Be Easily Distracted by Irrelevant Context},
  author =       {Shi, Freda and Chen, Xinyun and Misra, Kanishka and Scales, Nathan and Dohan, David and Chi, Ed H. and Sch\"{a}rli, Nathanael and Zhou, Denny},
  booktitle = 	 {Proceedings of the 40th International Conference on Machine Learning},
  pages = 	 {31210--31227},
  year = 	 {2023},
  editor = 	 {Krause, Andreas and Brunskill, Emma and Cho, Kyunghyun and Engelhardt, Barbara and Sabato, Sivan and Scarlett, Jonathan},
  volume = 	 {202},
  series = 	 {Proceedings of Machine Learning Research},
  month = 	 {23--29 Jul},
  publisher =    {PMLR},
  url = 	 {https://proceedings.mlr.press/v202/shi23a.html}
}

@misc{yuan_scaling_2023,
	title = {Scaling {Relationship} on {Learning} {Mathematical} {Reasoning} with {Large} {Language} {Models}},
	url = {http://arxiv.org/abs/2308.01825},
	urldate = {2024-04-13},
	publisher = {arXiv},
	author = {Yuan, Zheng and Yuan, Hongyi and Li, Chengpeng and Dong, Guanting and Lu, Keming and Tan, Chuanqi and Zhou, Chang and Zhou, Jingren},
	month = sep,
	year = {2023},
	note = {arXiv:2308.01825 [cs]},
}

@misc{mishra_lila_2023,
	title = {Lila: {A} {Unified} {Benchmark} for {Mathematical} {Reasoning}},
	shorttitle = {Lila},
	url = {http://arxiv.org/abs/2210.17517},
	doi = {10.48550/arXiv.2210.17517},
	urldate = {2024-04-13},
	publisher = {arXiv},
	author = {Mishra, Swaroop and Finlayson, Matthew and Lu, Pan and Tang, Leonard and Welleck, Sean and Baral, Chitta and Rajpurohit, Tanmay and Tafjord, Oyvind and Sabharwal, Ashish and Clark, Peter and Kalyan, Ashwin},
	month = mar,
	year = {2023},
	note = {arXiv:2210.17517 [cs]},
}

@article{toshniwal2024openmath,
  title   = {OpenMathInstruct-1: A 1.8 Million Math Instruction Tuning Dataset},
  author  = {Shubham Toshniwal and Ivan Moshkov and Sean Narenthiran and Daria Gitman and Fei Jia and Igor Gitman},
  year    = {2024},
  journal = {arXiv preprint arXiv: Arxiv-2402.10176}
}

@article{wu_automatic_2022,
	title = {Automatic {Math} {Word} {Problem} {Generation} {With} {Topic}-{Expression} {Co}-{Attention} {Mechanism} and {Reinforcement} {Learning}},
	volume = {30},
	copyright = {https://ieeexplore.ieee.org/Xplorehelp/downloads/license-information/IEEE.html},
	issn = {2329-9290, 2329-9304},
	url = {https://ieeexplore.ieee.org/document/9723526/},
	doi = {10.1109/TASLP.2022.3155284},
	language = {en},
	urldate = {2024-05-23},
	journal = {IEEE/ACM Transactions on Audio, Speech, and Language Processing},
	author = {Wu, Qinzhuo and Zhang, Qi and Huang, Xuanjing},
	year = {2022},
	pages = {1061--1072},
}

@article{walkington_how_2018,
	title = {How readability factors are differentially associated with performance for students of different backgrounds when solving mathematics word problems},
	volume = {55},
	issn = {1935-1011},
	doi = {10.3102/0002831217737028},
	number = {2},
	journal = {American Educational Research Journal},
	author = {Walkington, Candace and Clinton, Virginia and Shivraj, Pooja},
	year = {2018},
	note = {Place: US
Publisher: Sage Publications},
	pages = {362--414},
}

@techreport{baker_results_2020,
	title = {The {Results} of {Implementing} {Zone} of {Proximal} {Development} on {Learning} {Outcomes}},
	url = {https://eric.ed.gov/?id=ED608058},
	language = {en},
	urldate = {2024-07-08},
	institution = {International Educational Data Mining Society},
	author = {Baker, Ryan and Ma, Wei and Zhao, Yuxin and Wang, Shengni and Ma, Zhenjun},
	month = jul,
	year = {2020},
	note = {ERIC Number: ED608058},
}

@article{nasir_court_2008,
	title = {From the {Court} to the {Classroom}: {Opportunities} for {Engagement}, {Learning}, and {Identity} in {Basketball} and {Classroom} {Mathematics}},
	volume = {17},
	issn = {1050-8406},
	shorttitle = {From the {Court} to the {Classroom}},
	url = {https://doi.org/10.1080/10508400801986108},
	doi = {10.1080/10508400801986108},
	number = {2},
	urldate = {2024-07-16},
	journal = {Journal of the Learning Sciences},
	author = {Nasir, Na'ilah Suad and Hand, Victoria},
	month = apr,
	year = {2008},
	note = {Publisher: Routledge
\_eprint: https://doi.org/10.1080/10508400801986108},
	pages = {143--179},
}

@article{pinkard_equitable_2020,
	title = {Equitable approaches: opportunities for computational thinking with emphasis on creative production and connections to community},
	volume = {28},
	issn = {1049-4820},
	shorttitle = {Equitable approaches},
	url = {https://doi.org/10.1080/10494820.2019.1636070},
	doi = {10.1080/10494820.2019.1636070},
	number = {3},
	urldate = {2024-07-16},
	journal = {Interactive Learning Environments},
	author = {Pinkard, Nichole and Martin, C. K. and Erete, S.},
	month = apr,
	year = {2020},
	note = {Publisher: Routledge
\_eprint: https://doi.org/10.1080/10494820.2019.1636070},
	pages = {347--361},
}

@misc{openai_introducing_2025,
	title = {Introducing {GPT}-4.5},
	url = {https://openai.com/index/introducing-gpt-4-5/},
	language = {en-US},
	urldate = {2025-04-12},
	author = {OpenAI},
	month = feb,
	year = {2025},
}

@misc{openai_gpt-4o_2024,
	title = {{GPT}-4o {System} {Card}},
	url = {http://arxiv.org/abs/2410.21276},
	doi = {10.48550/arXiv.2410.21276},
	urldate = {2025-04-12},
	publisher = {arXiv},
	author = {OpenAI and Hurst, Aaron and Lerer, Adam and Goucher, Adam P. and Perelman, Adam and Ramesh, Aditya and Clark, Aidan and Ostrow, A. J. and Welihinda, Akila and Hayes, Alan and Radford, Alec and Mądry, Aleksander and Baker-Whitcomb, Alex and Beutel, Alex and Borzunov, Alex and Carney, Alex and Chow, Alex and Kirillov, Alex and Nichol, Alex and Paino, Alex and Renzin, Alex and Passos, Alex Tachard and Kirillov, Alexander and Christakis, Alexi and Conneau, Alexis and Kamali, Ali and Jabri, Allan and Moyer, Allison and Tam, Allison and Crookes, Amadou and Tootoochian, Amin and Tootoonchian, Amin and Kumar, Ananya and Vallone, Andrea and Karpathy, Andrej and Braunstein, Andrew and Cann, Andrew and Codispoti, Andrew and Galu, Andrew and Kondrich, Andrew and Tulloch, Andrew and Mishchenko, Andrey and Baek, Angela and Jiang, Angela and Pelisse, Antoine and Woodford, Antonia and Gosalia, Anuj and Dhar, Arka and Pantuliano, Ashley and Nayak, Avi and Oliver, Avital and Zoph, Barret and Ghorbani, Behrooz and Leimberger, Ben and Rossen, Ben and Sokolowsky, Ben and Wang, Ben and Zweig, Benjamin and Hoover, Beth and Samic, Blake and McGrew, Bob and Spero, Bobby and Giertler, Bogo and Cheng, Bowen and Lightcap, Brad and Walkin, Brandon and Quinn, Brendan and Guarraci, Brian and Hsu, Brian and Kellogg, Bright and Eastman, Brydon and Lugaresi, Camillo and Wainwright, Carroll and Bassin, Cary and Hudson, Cary and Chu, Casey and Nelson, Chad and Li, Chak and Shern, Chan Jun and Conger, Channing and Barette, Charlotte and Voss, Chelsea and Ding, Chen and Lu, Cheng and Zhang, Chong and Beaumont, Chris and Hallacy, Chris and Koch, Chris and Gibson, Christian and Kim, Christina and Choi, Christine and McLeavey, Christine and Hesse, Christopher and Fischer, Claudia and Winter, Clemens and Czarnecki, Coley and Jarvis, Colin and Wei, Colin and Koumouzelis, Constantin and Sherburn, Dane and Kappler, Daniel and Levin, Daniel and Levy, Daniel and Carr, David and Farhi, David and Mely, David and Robinson, David and Sasaki, David and Jin, Denny and Valladares, Dev and Tsipras, Dimitris and Li, Doug and Nguyen, Duc Phong and Findlay, Duncan and Oiwoh, Edede and Wong, Edmund and Asdar, Ehsan and Proehl, Elizabeth and Yang, Elizabeth and Antonow, Eric and Kramer, Eric and Peterson, Eric and Sigler, Eric and Wallace, Eric and Brevdo, Eugene and Mays, Evan and Khorasani, Farzad and Such, Felipe Petroski and Raso, Filippo and Zhang, Francis and Lohmann, Fred von and Sulit, Freddie and Goh, Gabriel and Oden, Gene and Salmon, Geoff and Starace, Giulio and Brockman, Greg and Salman, Hadi and Bao, Haiming and Hu, Haitang and Wong, Hannah and Wang, Haoyu and Schmidt, Heather and Whitney, Heather and Jun, Heewoo and Kirchner, Hendrik and Pinto, Henrique Ponde de Oliveira and Ren, Hongyu and Chang, Huiwen and Chung, Hyung Won and Kivlichan, Ian and O'Connell, Ian and O'Connell, Ian and Osband, Ian and Silber, Ian and Sohl, Ian and Okuyucu, Ibrahim and Lan, Ikai and Kostrikov, Ilya and Sutskever, Ilya and Kanitscheider, Ingmar and Gulrajani, Ishaan and Coxon, Jacob and Menick, Jacob and Pachocki, Jakub and Aung, James and Betker, James and Crooks, James and Lennon, James and Kiros, Jamie and Leike, Jan and Park, Jane and Kwon, Jason and Phang, Jason and Teplitz, Jason and Wei, Jason and Wolfe, Jason and Chen, Jay and Harris, Jeff and Varavva, Jenia and Lee, Jessica Gan and Shieh, Jessica and Lin, Ji and Yu, Jiahui and Weng, Jiayi and Tang, Jie and Yu, Jieqi and Jang, Joanne and Candela, Joaquin Quinonero and Beutler, Joe and Landers, Joe and Parish, Joel and Heidecke, Johannes and Schulman, John and Lachman, Jonathan and McKay, Jonathan and Uesato, Jonathan and Ward, Jonathan and Kim, Jong Wook and Huizinga, Joost and Sitkin, Jordan and Kraaijeveld, Jos and Gross, Josh and Kaplan, Josh and Snyder, Josh and Achiam, Joshua and Jiao, Joy and Lee, Joyce and Zhuang, Juntang and Harriman, Justyn and Fricke, Kai and Hayashi, Kai and Singhal, Karan and Shi, Katy and Karthik, Kavin and Wood, Kayla and Rimbach, Kendra and Hsu, Kenny and Nguyen, Kenny and Gu-Lemberg, Keren and Button, Kevin and Liu, Kevin and Howe, Kiel and Muthukumar, Krithika and Luther, Kyle and Ahmad, Lama and Kai, Larry and Itow, Lauren and Workman, Lauren and Pathak, Leher and Chen, Leo and Jing, Li and Guy, Lia and Fedus, Liam and Zhou, Liang and Mamitsuka, Lien and Weng, Lilian and McCallum, Lindsay and Held, Lindsey and Ouyang, Long and Feuvrier, Louis and Zhang, Lu and Kondraciuk, Lukas and Kaiser, Lukasz and Hewitt, Luke and Metz, Luke and Doshi, Lyric and Aflak, Mada and Simens, Maddie and Boyd, Madelaine and Thompson, Madeleine and Dukhan, Marat and Chen, Mark and Gray, Mark and Hudnall, Mark and Zhang, Marvin and Aljubeh, Marwan and Litwin, Mateusz and Zeng, Matthew and Johnson, Max and Shetty, Maya and Gupta, Mayank and Shah, Meghan and Yatbaz, Mehmet and Yang, Meng Jia and Zhong, Mengchao and Glaese, Mia and Chen, Mianna and Janner, Michael and Lampe, Michael and Petrov, Michael and Wu, Michael and Wang, Michele and Fradin, Michelle and Pokrass, Michelle and Castro, Miguel and Castro, Miguel Oom Temudo de and Pavlov, Mikhail and Brundage, Miles and Wang, Miles and Khan, Minal and Murati, Mira and Bavarian, Mo and Lin, Molly and Yesildal, Murat and Soto, Nacho and Gimelshein, Natalia and Cone, Natalie and Staudacher, Natalie and Summers, Natalie and LaFontaine, Natan and Chowdhury, Neil and Ryder, Nick and Stathas, Nick and Turley, Nick and Tezak, Nik and Felix, Niko and Kudige, Nithanth and Keskar, Nitish and Deutsch, Noah and Bundick, Noel and Puckett, Nora and Nachum, Ofir and Okelola, Ola and Boiko, Oleg and Murk, Oleg and Jaffe, Oliver and Watkins, Olivia and Godement, Olivier and Campbell-Moore, Owen and Chao, Patrick and McMillan, Paul and Belov, Pavel and Su, Peng and Bak, Peter and Bakkum, Peter and Deng, Peter and Dolan, Peter and Hoeschele, Peter and Welinder, Peter and Tillet, Phil and Pronin, Philip and Tillet, Philippe and Dhariwal, Prafulla and Yuan, Qiming and Dias, Rachel and Lim, Rachel and Arora, Rahul and Troll, Rajan and Lin, Randall and Lopes, Rapha Gontijo and Puri, Raul and Miyara, Reah and Leike, Reimar and Gaubert, Renaud and Zamani, Reza and Wang, Ricky and Donnelly, Rob and Honsby, Rob and Smith, Rocky and Sahai, Rohan and Ramchandani, Rohit and Huet, Romain and Carmichael, Rory and Zellers, Rowan and Chen, Roy and Chen, Ruby and Nigmatullin, Ruslan and Cheu, Ryan and Jain, Saachi and Altman, Sam and Schoenholz, Sam and Toizer, Sam and Miserendino, Samuel and Agarwal, Sandhini and Culver, Sara and Ethersmith, Scott and Gray, Scott and Grove, Sean and Metzger, Sean and Hermani, Shamez and Jain, Shantanu and Zhao, Shengjia and Wu, Sherwin and Jomoto, Shino and Wu, Shirong and Shuaiqi and Xia and Phene, Sonia and Papay, Spencer and Narayanan, Srinivas and Coffey, Steve and Lee, Steve and Hall, Stewart and Balaji, Suchir and Broda, Tal and Stramer, Tal and Xu, Tao and Gogineni, Tarun and Christianson, Taya and Sanders, Ted and Patwardhan, Tejal and Cunninghman, Thomas and Degry, Thomas and Dimson, Thomas and Raoux, Thomas and Shadwell, Thomas and Zheng, Tianhao and Underwood, Todd and Markov, Todor and Sherbakov, Toki and Rubin, Tom and Stasi, Tom and Kaftan, Tomer and Heywood, Tristan and Peterson, Troy and Walters, Tyce and Eloundou, Tyna and Qi, Valerie and Moeller, Veit and Monaco, Vinnie and Kuo, Vishal and Fomenko, Vlad and Chang, Wayne and Zheng, Weiyi and Zhou, Wenda and Manassra, Wesam and Sheu, Will and Zaremba, Wojciech and Patil, Yash and Qian, Yilei and Kim, Yongjik and Cheng, Youlong and Zhang, Yu and He, Yuchen and Zhang, Yuchen and Jin, Yujia and Dai, Yunxing and Malkov, Yury},
	month = oct,
	year = {2024},
	note = {arXiv:2410.21276 [cs]},
}

@misc{vadoe_sol_nodate,
	title = {{SOL} {Practice} {Items} ({All} {Subjects}) {\textbar} {Virginia} {Department} of {Education}},
	url = {https://www.doe.virginia.gov/teaching-learning-assessment/student-assessment/sol-practice-items-all-subjects},
	language = {en},
	urldate = {2023-03-22},
	author = {VDOE},
}

@misc{tang_mathscale_2024,
	title = {{MathScale}: {Scaling} {Instruction} {Tuning} for {Mathematical} {Reasoning}},
	shorttitle = {{MathScale}},
	url = {http://arxiv.org/abs/2403.02884},
	doi = {10.48550/arXiv.2403.02884},
	urldate = {2024-03-06},
	publisher = {arXiv},
	author = {Tang, Zhengyang and Zhang, Xingxing and Wang, Benyou and Wei, Furu},
	month = mar,
	year = {2024},
	note = {arXiv:2403.02884 [cs]},
}

@misc{openai_gpt-5_2025,
	title = {{GPT}-5 {System} {Card}},
	url = {https://openai.com/index/gpt-5-system-card/},
	language = {en-US},
	urldate = {2025-12-15},
	author = {OpenAI},
	month = dec,
	year = {2025},
}

@misc{li_synthetic_2024,
	title = {Synthetic {Data} ({Almost}) from {Scratch}: {Generalized} {Instruction} {Tuning} for {Language} {Models}},
	shorttitle = {Synthetic {Data} ({Almost}) from {Scratch}},
	url = {http://arxiv.org/abs/2402.13064},
	doi = {10.48550/arXiv.2402.13064},
	urldate = {2025-12-15},
	publisher = {arXiv},
	author = {Li, Haoran and Dong, Qingxiu and Tang, Zhengyang and Wang, Chaojun and Zhang, Xingxing and Huang, Haoyang and Huang, Shaohan and Huang, Xiaolong and Huang, Zeqiang and Zhang, Dongdong and Gu, Yuxian and Cheng, Xin and Wang, Xun and Chen, Si-Qing and Dong, Li and Lu, Wei and Sui, Zhifang and Wang, Benyou and Lam, Wai and Wei, Furu},
	month = feb,
	year = {2024},
	note = {arXiv:2402.13064 [cs]},
}

@misc{kuncl_overworked_2024,
	title = {Overworked and {Undervalued}: {Retaining} {Top} {Educators}},
	shorttitle = {Overworked and {Undervalued}},
	url = {https://www.gallup.com/education/609422/overworked-undervalued-retaining-top-educators.aspx},
	language = {en},
	urldate = {2024-07-08},
	journal = {Gallup.com},
	author = {Kuncl, Liz and Christensen, Kyle},
	month = feb,
	year = {2024},
	note = {Section: Education},
}

@misc{common_core_state_standards_initiative_common_nodate,
	title = {Common {Core} {State} {Standards} {Initiative} – {Preparing} {America}'s {Students} for {College} \& {Career}},
	url = {https://corestandards.org/},
	language = {en-US},
	urldate = {2025-04-07},
	author = {Common Core State Standards Initiative},
}

@misc{meta_llama_nodate,
	title = {Llama 3.3 {Model} {Cards} and {Prompt} formats},
	url = {https://www.llama.com/docs/model-cards-and-prompt-formats/llama3_3/},
	language = {en},
	urldate = {2025-04-08},
	author = {Meta},
}

@misc{ethayarajh_kto_2024,
	title = {{KTO}: {Model} {Alignment} as {Prospect} {Theoretic} {Optimization}},
	shorttitle = {{KTO}},
	url = {http://arxiv.org/abs/2402.01306},
	doi = {10.48550/arXiv.2402.01306},
	urldate = {2024-05-27},
	publisher = {arXiv},
	author = {Ethayarajh, Kawin and Xu, Winnie and Muennighoff, Niklas and Jurafsky, Dan and Kiela, Douwe},
	month = feb,
	year = {2024},
	note = {arXiv:2402.01306 [cs]},
}

@misc{team_gemma_2025,
	title = {Gemma 3 {Technical} {Report}},
	url = {http://arxiv.org/abs/2503.19786},
	doi = {10.48550/arXiv.2503.19786},
	urldate = {2025-04-09},
	publisher = {arXiv},
	author = {Team, Gemma and Kamath, Aishwarya and Ferret, Johan and Pathak, Shreya and Vieillard, Nino and Merhej, Ramona and Perrin, Sarah and Matejovicova, Tatiana and Ramé, Alexandre and Rivière, Morgane and Rouillard, Louis and Mesnard, Thomas and Cideron, Geoffrey and Grill, Jean-bastien and Ramos, Sabela and Yvinec, Edouard and Casbon, Michelle and Pot, Etienne and Penchev, Ivo and Liu, Gaël and Visin, Francesco and Kenealy, Kathleen and Beyer, Lucas and Zhai, Xiaohai and Tsitsulin, Anton and Busa-Fekete, Robert and Feng, Alex and Sachdeva, Noveen and Coleman, Benjamin and Gao, Yi and Mustafa, Basil and Barr, Iain and Parisotto, Emilio and Tian, David and Eyal, Matan and Cherry, Colin and Peter, Jan-Thorsten and Sinopalnikov, Danila and Bhupatiraju, Surya and Agarwal, Rishabh and Kazemi, Mehran and Malkin, Dan and Kumar, Ravin and Vilar, David and Brusilovsky, Idan and Luo, Jiaming and Steiner, Andreas and Friesen, Abe and Sharma, Abhanshu and Sharma, Abheesht and Gilady, Adi Mayrav and Goedeckemeyer, Adrian and Saade, Alaa and Feng, Alex and Kolesnikov, Alexander and Bendebury, Alexei and Abdagic, Alvin and Vadi, Amit and György, András and Pinto, André Susano and Das, Anil and Bapna, Ankur and Miech, Antoine and Yang, Antoine and Paterson, Antonia and Shenoy, Ashish and Chakrabarti, Ayan and Piot, Bilal and Wu, Bo and Shahriari, Bobak and Petrini, Bryce and Chen, Charlie and Lan, Charline Le and Choquette-Choo, Christopher A. and Carey, C. J. and Brick, Cormac and Deutsch, Daniel and Eisenbud, Danielle and Cattle, Dee and Cheng, Derek and Paparas, Dimitris and Sreepathihalli, Divyashree Shivakumar and Reid, Doug and Tran, Dustin and Zelle, Dustin and Noland, Eric and Huizenga, Erwin and Kharitonov, Eugene and Liu, Frederick and Amirkhanyan, Gagik and Cameron, Glenn and Hashemi, Hadi and Klimczak-Plucińska, Hanna and Singh, Harman and Mehta, Harsh and Lehri, Harshal Tushar and Hazimeh, Hussein and Ballantyne, Ian and Szpektor, Idan and Nardini, Ivan and Pouget-Abadie, Jean and Chan, Jetha and Stanton, Joe and Wieting, John and Lai, Jonathan and Orbay, Jordi and Fernandez, Joseph and Newlan, Josh and Ji, Ju-yeong and Singh, Jyotinder and Black, Kat and Yu, Kathy and Hui, Kevin and Vodrahalli, Kiran and Greff, Klaus and Qiu, Linhai and Valentine, Marcella and Coelho, Marina and Ritter, Marvin and Hoffman, Matt and Watson, Matthew and Chaturvedi, Mayank and Moynihan, Michael and Ma, Min and Babar, Nabila and Noy, Natasha and Byrd, Nathan and Roy, Nick and Momchev, Nikola and Chauhan, Nilay and Sachdeva, Noveen and Bunyan, Oskar and Botarda, Pankil and Caron, Paul and Rubenstein, Paul Kishan and Culliton, Phil and Schmid, Philipp and Sessa, Pier Giuseppe and Xu, Pingmei and Stanczyk, Piotr and Tafti, Pouya and Shivanna, Rakesh and Wu, Renjie and Pan, Renke and Rokni, Reza and Willoughby, Rob and Vallu, Rohith and Mullins, Ryan and Jerome, Sammy and Smoot, Sara and Girgin, Sertan and Iqbal, Shariq and Reddy, Shashir and Sheth, Shruti and Põder, Siim and Bhatnagar, Sijal and Panyam, Sindhu Raghuram and Eiger, Sivan and Zhang, Susan and Liu, Tianqi and Yacovone, Trevor and Liechty, Tyler and Kalra, Uday and Evci, Utku and Misra, Vedant and Roseberry, Vincent and Feinberg, Vlad and Kolesnikov, Vlad and Han, Woohyun and Kwon, Woosuk and Chen, Xi and Chow, Yinlam and Zhu, Yuvein and Wei, Zichuan and Egyed, Zoltan and Cotruta, Victor and Giang, Minh and Kirk, Phoebe and Rao, Anand and Black, Kat and Babar, Nabila and Lo, Jessica and Moreira, Erica and Martins, Luiz Gustavo and Sanseviero, Omar and Gonzalez, Lucas and Gleicher, Zach and Warkentin, Tris and Mirrokni, Vahab and Senter, Evan and Collins, Eli and Barral, Joelle and Ghahramani, Zoubin and Hadsell, Raia and Matias, Yossi and Sculley, D. and Petrov, Slav and Fiedel, Noah and Shazeer, Noam and Vinyals, Oriol and Dean, Jeff and Hassabis, Demis and Kavukcuoglu, Koray and Farabet, Clement and Buchatskaya, Elena and Alayrac, Jean-Baptiste and Anil, Rohan and Dmitry and Lepikhin and Borgeaud, Sebastian and Bachem, Olivier and Joulin, Armand and Andreev, Alek and Hardin, Cassidy and Dadashi, Robert and Hussenot, Léonard},
	month = mar,
	year = {2025},
	note = {arXiv:2503.19786 [cs]},
}

@misc{ariyarathne_elementary_2025,
	title = {Elementary {Math} {Word} {Problem} {Generation} using {Large} {Language} {Models}},
	url = {http://arxiv.org/abs/2506.05950},
	doi = {10.48550/arXiv.2506.05950},
	urldate = {2025-06-13},
	publisher = {arXiv},
	author = {Ariyarathne, Nimesh and Bandara, Harshani and Heshan, Yasith and Gamage, Omega and Ranathunga, Surangika and Nayanajith, Dilan and Sivapalan, Yutharsan and Lihinikaduarachchi, Gayathri and Vihidun, Tharoosha and Chandirakumar, Meenambika and Premakumar, Sanujen and Gathsara, Sanjula},
	month = jun,
	year = {2025},
	note = {arXiv:2506.05950 [cs]},
}

@article{crossley_assessing_2008,
	title = {Assessing {Text} {Readability} {Using} {Cognitively} {Based} {Indices}},
	volume = {42},
	copyright = {2008 TESOL International Association},
	issn = {1545-7249},
	url = {https://onlinelibrary.wiley.com/doi/abs/10.1002/j.1545-7249.2008.tb00142.x},
	doi = {10.1002/j.1545-7249.2008.tb00142.x},
	language = {en},
	number = {3},
	urldate = {2025-08-26},
	journal = {TESOL Quarterly},
	author = {Crossley, Scott A. and Greenfield, Jerry and McNAMARA, Danielle S.},
	year = {2008},
	note = {\_eprint: https://onlinelibrary.wiley.com/doi/pdf/10.1002/j.1545-7249.2008.tb00142.x},
	pages = {475--493},
}

@article{mc_laughlin_smog_1969,
	title = {{SMOG} {Grading}-a {New} {Readability} {Formula}},
	volume = {12},
	issn = {0022-4103},
	url = {https://www.jstor.org/stable/40011226},
	number = {8},
	urldate = {2025-08-26},
	journal = {Journal of Reading},
	author = {Mc Laughlin, G. Harry},
	year = {1969},
	note = {Publisher: [Wiley, International Reading Association]},
	pages = {639--646},
}

@article{crossley_moving_2019,
	title = {Moving beyond classic readability formulas: new methods and new models},
	volume = {42},
	copyright = {© 2019 UKLA},
	issn = {1467-9817},
	shorttitle = {Moving beyond classic readability formulas},
	url = {https://onlinelibrary.wiley.com/doi/abs/10.1111/1467-9817.12283},
	doi = {10.1111/1467-9817.12283},
	language = {en},
	number = {3-4},
	urldate = {2024-09-23},
	journal = {Journal of Research in Reading},
	author = {Crossley, Scott A. and Skalicky, Stephen and Dascalu, Mihai},
	year = {2019},
	note = {\_eprint: https://onlinelibrary.wiley.com/doi/pdf/10.1111/1467-9817.12283},
	pages = {541--561},
}

@misc{warner_smarter_2024,
	title = {Smarter, {Better}, {Faster}, {Longer}: {A} {Modern} {Bidirectional} {Encoder} for {Fast}, {Memory} {Efficient}, and {Long} {Context} {Finetuning} and {Inference}},
	shorttitle = {Smarter, {Better}, {Faster}, {Longer}},
	url = {http://arxiv.org/abs/2412.13663},
	doi = {10.48550/arXiv.2412.13663},
	urldate = {2025-09-08},
	publisher = {arXiv},
	author = {Warner, Benjamin and Chaffin, Antoine and Clavié, Benjamin and Weller, Orion and Hallström, Oskar and Taghadouini, Said and Gallagher, Alexis and Biswas, Raja and Ladhak, Faisal and Aarsen, Tom and Cooper, Nathan and Adams, Griffin and Howard, Jeremy and Poli, Iacopo},
	month = dec,
	year = {2024},
	note = {arXiv:2412.13663 [cs]},
}

@misc{yang_qwen3_2025,
	title = {Qwen3 {Technical} {Report}},
	url = {http://arxiv.org/abs/2505.09388},
	doi = {10.48550/arXiv.2505.09388},
	urldate = {2025-09-08},
	publisher = {arXiv},
	author = {Yang, An and Li, Anfeng and Yang, Baosong and Zhang, Beichen and Hui, Binyuan and Zheng, Bo and Yu, Bowen and Gao, Chang and Huang, Chengen and Lv, Chenxu and Zheng, Chujie and Liu, Dayiheng and Zhou, Fan and Huang, Fei and Hu, Feng and Ge, Hao and Wei, Haoran and Lin, Huan and Tang, Jialong and Yang, Jian and Tu, Jianhong and Zhang, Jianwei and Yang, Jianxin and Yang, Jiaxi and Zhou, Jing and Zhou, Jingren and Lin, Junyang and Dang, Kai and Bao, Keqin and Yang, Kexin and Yu, Le and Deng, Lianghao and Li, Mei and Xue, Mingfeng and Li, Mingze and Zhang, Pei and Wang, Peng and Zhu, Qin and Men, Rui and Gao, Ruize and Liu, Shixuan and Luo, Shuang and Li, Tianhao and Tang, Tianyi and Yin, Wenbiao and Ren, Xingzhang and Wang, Xinyu and Zhang, Xinyu and Ren, Xuancheng and Fan, Yang and Su, Yang and Zhang, Yichang and Zhang, Yinger and Wan, Yu and Liu, Yuqiong and Wang, Zekun and Cui, Zeyu and Zhang, Zhenru and Zhou, Zhipeng and Qiu, Zihan},
	month = may,
	year = {2025},
	note = {arXiv:2505.09388 [cs]},
}

\appendix
\section{Model Development Pipeline} \label{model_pipeline}
\begin{figure*}
    \centering
    \includegraphics[width=1\linewidth]{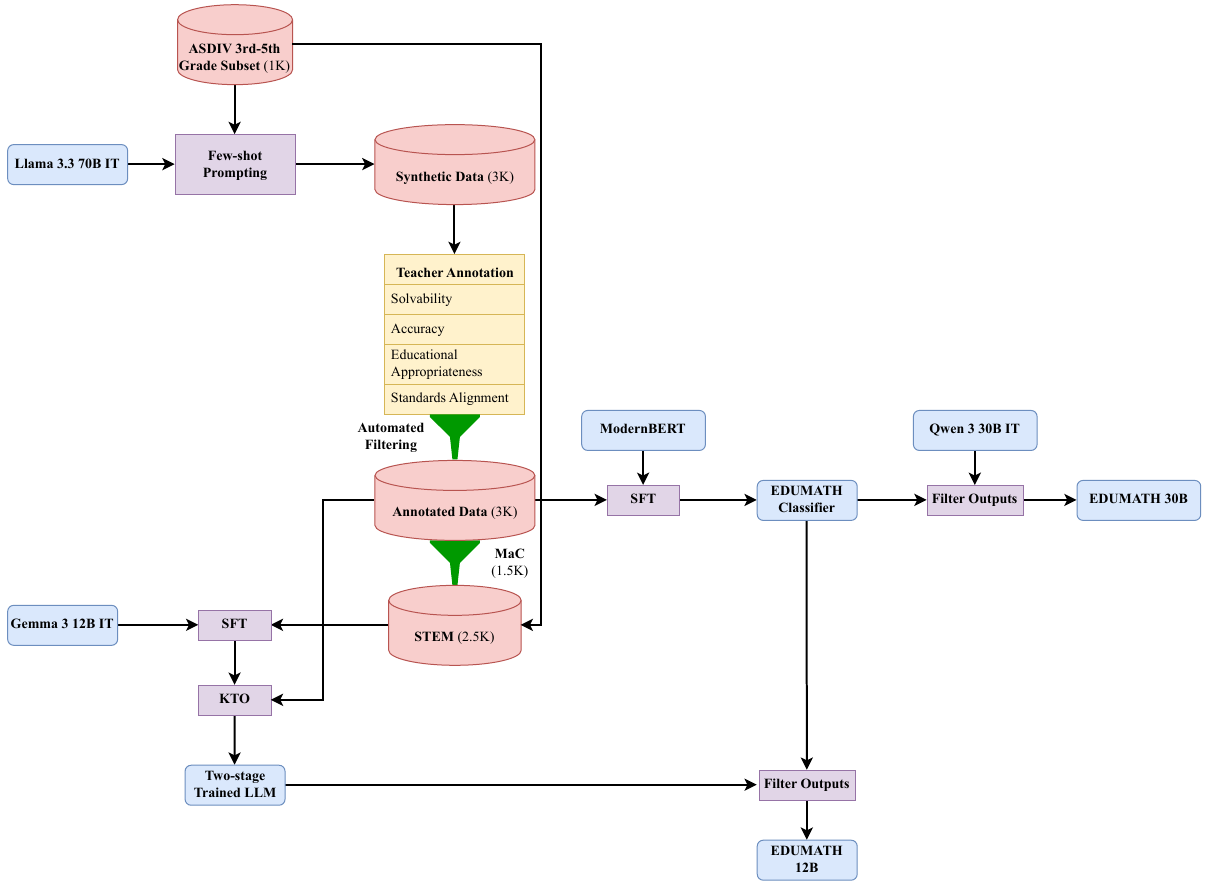}
    \caption{Visual of STEM dataset and EDUMATH models' development process. Blue rounded rectangles denote LLMs, red cylinders denote data sources, purple rectangles denote model interventions/training processes, green funnels denote a dataset filtering process, directional arrows represent inputs/outputs, and MaC denotes meets all criteria.}
    \label{fig:model_pipeline}
\end{figure*}

Figure \ref{fig:model_pipeline} visually displays the process of developing the STEM dataset and EDUMATH models described in Section \ref{methods}.
\section{Annotator Directions and Demographics} \label{EDUMATH:annotator_directions}
Figures \ref{fig:EDUMATH_solvability_directions}, \ref{fig:EDUMATH_accuracy_directions}, \ref{fig:EDUMATH_appropriateness_directions}, and \ref{fig:EDUMATH_standards_alignment_directions} display the directions presented to annotators for annotating MWPs for solvability, accuracy, educational appropriateness, and standards alignment, respectively. If an annotator answered "No" to solvability or any other criteria, they were instructed to select "NA" for the remaining criteria given that a failure to meet one of the criteria automatically means a MWP does not MaC. This saved the annotators time, as they did not need to unnecessarily annotate the remaining criteria. Table \ref{tab:annotator_demographics} displays annotator demographics and average time spent per MWP. Our human annotation study was approved by the institutional review board at our host university.
\begin{figure*}
  \centering
  \begin{tcolorbox}[
      colback=gray!5,
      colframe=black,
      width=0.9\linewidth,
      sharp corners,
      boxrule=0.6pt,
      left=4pt, right=4pt, top=4pt, bottom=4pt
    ]
    Is this question solvable? A solvable question means that it can be solved with the information present and does not contain a mathematical scenario that is impossible (e.g., giving away more money or items than you have). \\[0.5em]
- Yes \\[0.5em]
- No
  \end{tcolorbox}
  \caption{Annotator directions and response options for annotating MWPs for solvability.}
  \label{fig:EDUMATH_solvability_directions}
\end{figure*}

\begin{figure*}
  \centering
  \begin{tcolorbox}[
      colback=gray!5,
      colframe=black,
      width=0.9\linewidth,
      sharp corners,
      boxrule=0.6pt,
      left=4pt, right=4pt, top=4pt, bottom=4pt
    ]
Is the solution for this question correct? If the answer is correct but the reasoning is wrong, does not make sense, and/or is too complicated for a student/teacher to read, answer “No.”\\[0.5em]

- Yes, the solution and reasoning are correct.\\[0.5em]

- No. The solution is incorrect.\\[0.5em]

- No. The solution is correct, but the reasoning contains an error, does not make sense, and/or is too complicated for a student/teacher to read.\\[0.5em]

- NA
  \end{tcolorbox}
  \caption{Annotator directions and response options for annotating MWPs for accuracy.}
  \label{fig:EDUMATH_accuracy_directions}
\end{figure*}
\begin{figure*}
  \centering
  \begin{tcolorbox}[
      colback=gray!5,
      colframe=black,
      width=0.9\linewidth,
      sharp corners,
      boxrule=0.6pt,
      left=4pt, right=4pt, top=4pt, bottom=4pt
    ]
Would you feel comfortable giving this question to a student in a 3\textsuperscript{rd}-5\textsuperscript{th} grade school setting? \\[0.5em]

- Yes: This question is understandable and appropriate for a school setting. \\[0.5em]

- No: This question contains material inappropriate for a school setting (e.g., language about harming someone). \\[0.5em]

- No: This question is strange, confusing, contains conflicting information, and/or is not based in reality (e.g., contains misinformation). \\[0.5em]

- No: This question requires no mathematical operations to solve because it gives the answer away.\\[0.5em]

- No: It is inappropriate for a different reason.\\[0.5em]

- NA

  \end{tcolorbox}
  \caption{Annotator directions and response options for annotating MWPs for educational appropriateness.}
  \label{fig:EDUMATH_appropriateness_directions}
\end{figure*}
\begin{figure*}
  \centering
  \begin{tcolorbox}[
      colback=gray!5,
      colframe=black,
      width=0.9\linewidth,
      sharp corners,
      boxrule=0.6pt,
      left=4pt, right=4pt, top=4pt, bottom=4pt
    ]
Does the question adequately address important elements from the pre-specified math topic(s)? \\[0.5em]

If more than one math topic is listed, then the question should incorporate important elements of EACH listed math topic. If one of the topics lists multiple mathematical operations like addition, subtraction, and division, it is okay if the question just addresses one of those operations. \\[0.5em]

Select the most appropriate option below. \\[0.5em]

- Yes: The question addresses important elements from the specified topic(s)\\[0.5em]

- No: The question is too hard for the given topic(s). \\[0.5em]

- No: The question does not address some important parts of the specified topic(s). \\[0.5em]

- No: The question does not address the specified topic(s) at all. \\[0.5em]

- No: The question requires additional math topics or operations that are not listed in the specified math topic(s). \\[0.5em]

- NA

  \end{tcolorbox}
  \caption{Annotator directions and response options for annotating MWPs for standards alignment.}
  \label{fig:EDUMATH_standards_alignment_directions}
\end{figure*}
\begin{table}[t]
    \centering
    \begin{tabular}{lc}
    \toprule
        \textbf{Variable}& \textbf{Mean}\\
        \hline
        Age& 39.3 (11.7)\\
        Time Taken Per MWP (in seconds)& 117.0 (75.0)\\ \midrule
        & \textbf{Percentage}\\ \midrule
        Female& 67.1\\
        Race or Ethnicity& \\
        \hspace{3mm} White& 73.6\\
        \hspace{3mm} Black& 12.0\\
        \hspace{3mm} Asian& 5.9\\
        \hspace{3mm} Mixed& 5.4\\
 \hspace{3mm} Other&2.8\\
        \hspace{3mm} Prefer not to say& 0.3\\ \bottomrule
    \end{tabular}
    \caption{Annotator demographics and average time spent per MWP. Standard deviations, where appropriate, are in parentheses.}
    \label{tab:annotator_demographics}
\end{table}
\section{Model Training Details and Hyperparameters} \label{EDUMATH_training_hyperparameters}
\subsection{EDUMATH 12B}
For the supervised finetuning (SFT) training stage, we trained Gemma 3 12B IT on the STEM dataset for 5 epochs using 4 A6000 GPUs, a 85/15 train/validation split, a learning rate of 1e-6, a batch size of 1, and a 10\% warm-up ratio. We saved the model every 1,500 training steps and selected the model with the lowest validation loss as our final model (this was the model after step 10,000). For the Kahneman-Tversky Optimization (KTO) training stage, we continued training our SFT model on a dataset of 4,039 rows that combined our annotated data and the ASDIV 3rd-5th grade subset. We trained on this dataset using 4 A100 80GB GPUs for 5 epochs with an 85/15 train/validation split, learning rate of 5e-6, batch size of 8, a 10\% warm-up ratio, and desirable/undesirable weights based on inverse class frequency. We saved the model every 500 steps and selected the model with the lowest validation loss as our final model, which was the final model checkpoint. 

\subsection{Text Classifier}
We trained our ModernBERT \cite{warner_smarter_2024} text classifier such that a prediction of 1 means a MWP does not MaC and a prediction of 0 means a MWP does MaC. The training dataset combined our annotation data and MWPs from the ASDIV 3rd-5th grade subset that addressed standards that were not already in our annotated dataset for a total of 3,664 rows. We combined the data in this way to ensure that all standards were represented in our training data without over-representing positive examples for standards that were already in the annotated dataset. Given that there were a higher share of MWPs that MaC after adding in the ASDIV 3rd-5th grade subset (39.9\% of rows did not MaC), we trained the model using a loss term weighted by the inverse class balance. We trained the model using an A6000 GPU for 10 epochs with an 80/10/10 training/validation/test split, learning rate of 1e-5, batch size of 8, and a 10\% warm-up ratio, selecting the model checkpoint with the lowest validation loss as our final model (this was the model checkpoint after epoch 6). This model achieved a test accuracy of 79.0\% and AUC-ROC of .861. 

\subsection{Finetuning Ablation} \label{finetuning_ablation}
Table \ref{tab:finetuning_ablation_EDUMATH} shows an ablation for each EDUMATH 12B finetuning stage. Notably, each stage results in a statistically significant increase in MaC relative to the preceding stage, demonstrating the value of our annotated data for training standards-aligned educational MWP generators.
\begin{table}[t]
  \centering
  \begin{tabular}{l|c}
    \hline
    \textbf{Model}& \textbf{MaC}\\
    \hline
    Gemma 3 12B IT& 63.9 (1.5)\\
    \hspace{3mm} SFT& 76.2 (1.3)*\\
 \hspace{3mm}\hspace{3mm}KTO&81.0 (1.2)*\\
    \hspace{3mm}\hspace{3mm}\hspace{3mm}KTO + Classifier& 85.9 (1.2)*\\
    \hline
  \end{tabular}
  \caption{EDUMATH finetuning stages ablation. MaC denotes the share of 1,000 MWPs from each model that meet all criteria according to our Gemma 3 27B IT annotator. Standard errors are in parentheses and a * indicates a statistically significant difference between the current and previous stage at the p < 0.01 level.}
  \label{tab:finetuning_ablation_EDUMATH}
\end{table}

\section{Standard Counts in Labeled ASDIV Subset} \label{standards_counts}
\begin{figure}
    \centering
    \includegraphics[width=1\linewidth]{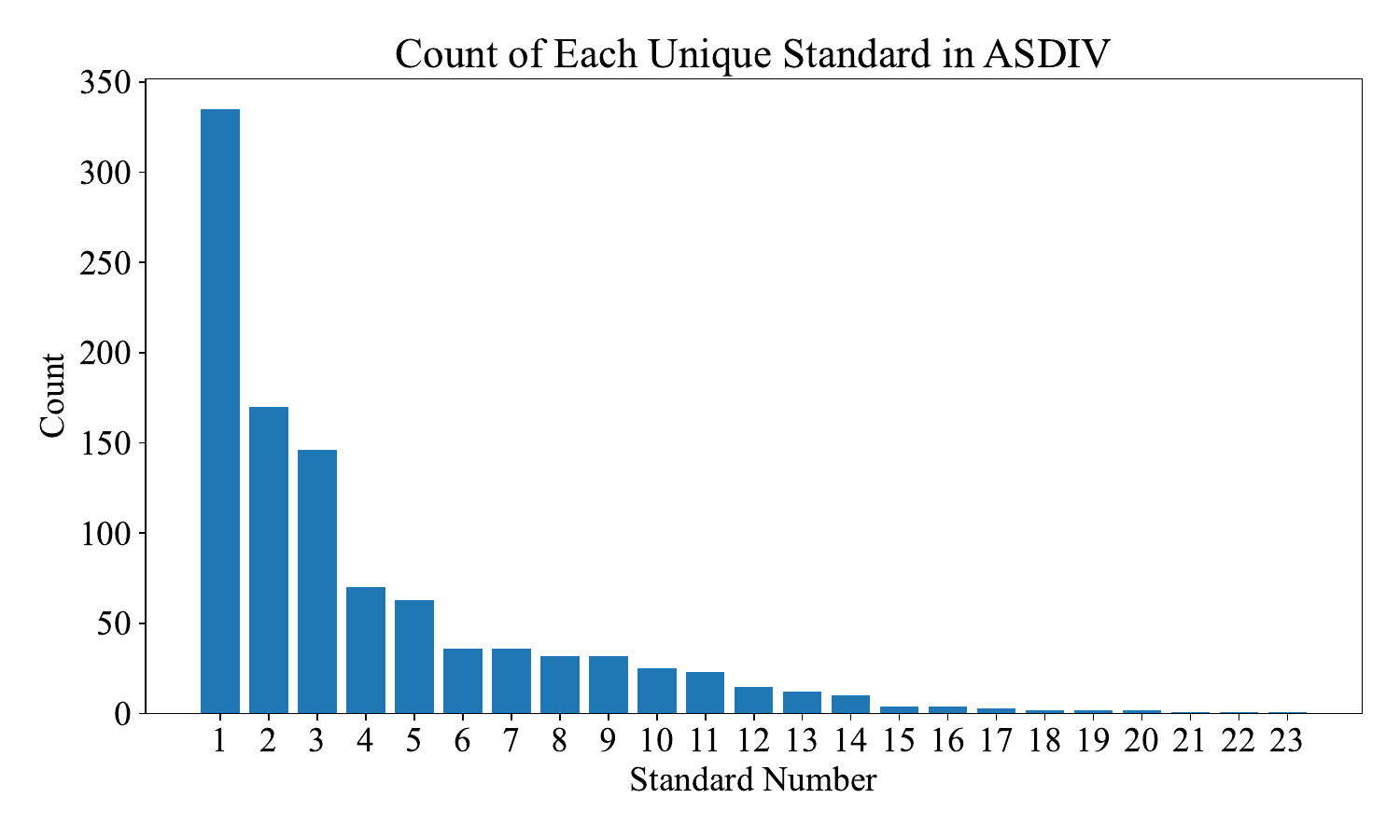}
    \caption{Count for each standard in the ASDIV 3rd-5th grade subset. A mapping of each standard number to its text is provided in the paper repo.}
    \label{fig:asdiv_standard_count}
\end{figure}

Figure \ref{fig:asdiv_standard_count} displays the count of MWPs from each standard included in the ASDIV 3rd-5th grade subset.
\begin{table*}[t]
\centering
\resizebox{\linewidth}{!}{
\small
\begin{tabular}{clcccc}
\toprule
& \textbf{Model}& \textbf{Solvability}&\textbf{Accuracy}& \textbf{Educational Appropriateness}&\textbf{Standards Alignment}\\ \midrule
\parbox[t]{2mm}{\multirow{3}{*}{\rotatebox[origin=c]{90}{API}}}&
GPT-4o& 5.6&  50.0& 0.0&44.4\\
 & GPT-4.1& 0.0& 61.1& 0.0& 38.9\\
& GPT-4.5& 5.0&  50.0& 0.0&45.0\\
\midrule
\parbox[t]{2mm}{\multirow{6}{*}{\rotatebox[origin=c]{90}{Public}}}&
Gemma 3 12B IT& 7.1& 43.1& 2.5&47.3\\
& Gemma 3 27B IT& 4.4& 48.8& 0.8&46.0\\
 & Qwen 3 30B IT& 1.5&  37.2& 0.8&60.5\\
 & Qwen 3 235B IT& 0.9&  38.6& 1.7&58.8\\
 & EDUMATH 12B (Ours)& 0.8& 31.7& 5.0&62.5\\
& EDUMATH 30B (Ours)& 5.5& 41.7& 0.0&52.8\\
\bottomrule
\end{tabular}
}
\caption{Proportion of overall errors from each evaluation criteria by model, or \textit{P(criteria failure | error)}.}
\label{tab:errors_by_criteria_EDUMATH}
\end{table*}
\begin{table*}[t]
\centering
\resizebox{.75\linewidth}{!}{
\small
\begin{tabular}{clcc}
\toprule
& \textbf{Model}& \textbf{Wrong Answer}&\textbf{\shortstack{Right Answer,\\ Poor or Confusing Reasoning}}\\ \midrule
\parbox[t]{2mm}{\multirow{3}{*}{\rotatebox[origin=c]{90}{API}}}&
GPT-4o& 33.3& 66.7\\
 & GPT-4.1& 45.5&54.5\\
& GPT-4.5& 50.0& 50.0\\
\midrule
\parbox[t]{2mm}{\multirow{6}{*}{\rotatebox[origin=c]{90}{Public}}}&
Gemma 3 12B IT& 45.9&54.1\\
& Gemma 3 27B IT& 42.3&57.7\\
 & Qwen 3 30B IT& 22.9& 77.1\\
 & Qwen 3 235B IT& 36.4& 63.6\\
 & EDUMATH 12B (Ours)& 44.7&55.3\\
& EDUMATH 30B (Ours)& 26.7&73.3\\
\bottomrule
\end{tabular}
}
\caption{Classification of accuracy errors for each model. See Figure \ref{fig:EDUMATH_accuracy_directions} for directions presented to our Gemma 3 27B IT automated annotator.}
\label{tab:accuracy_classification_EDUMATH}
\end{table*}
\begin{table*}[t]
\centering
\resizebox{.75\linewidth}{!}{
\small
\begin{tabular}{clcccc}
\toprule
& \textbf{Model}& \textbf{Strange}&\textbf{Harmful}&\textbf{No Ops}&\textbf{Other}\\ \midrule
\parbox[t]{2mm}{\multirow{3}{*}{\rotatebox[origin=c]{90}{API}}}&
GPT-4o& 0.0& 0.0& 0.0&0.0\\
 & GPT-4.1& 0.0& 0.0& 0.0&0.0\\
& GPT-4.5& 0.0& 0.0& 0.0&0.0\\
\midrule
\parbox[t]{2mm}{\multirow{6}{*}{\rotatebox[origin=c]{90}{Public}}}&
Gemma 3 12B IT& 88.9&0.0&11.1&0.0\\
& Gemma 3 27B IT& 50.0&0.0&50.0&0.0\\
 & Qwen 3 30B IT& 100.0& 0.0& 0.0&0.0\\
 & Qwen 3 235B IT& 100.0& 0.0& 0.0&0.0\\
 & EDUMATH 12B (Ours)& 100.0&0.0&0.0&0.0\\
& EDUMATH 30B (Ours)& 0.0&0.0&0.0&0.0\\
\bottomrule
\end{tabular}
}
\caption{Classification of appropriateness errors for each model. See Figure \ref{fig:EDUMATH_appropriateness_directions} for directions presented to our Gemma 3 27B IT automated annotator.}
\label{tab:appropriateness_classification_EDUMATH}
\end{table*}

\begin{table*}[t]
\centering
\resizebox{\linewidth}{!}{
\small
\begin{tabular}{clcccc}
\toprule
& \textbf{Model}& \textbf{Too Hard}&\textbf{\shortstack{Missing Important\\ Parts}}&\textbf{\shortstack{Does Not\\ Address Standard}} &\textbf{\shortstack{Requires\\Additional Topics}}\\ \midrule
\parbox[t]{2mm}{\multirow{3}{*}{\rotatebox[origin=c]{90}{API}}}&
GPT-4o& 0.0& 100.0&  0.0&0.0\\
 & GPT-4.1& 0.0& 71.4& 0.0&28.6\\
& GPT-4.5& 0.0& 77.8&  11.1&11.1\\
\midrule
\parbox[t]{2mm}{\multirow{6}{*}{\rotatebox[origin=c]{90}{Public}}}&
Gemma 3 12B IT& 5.8&61.0& 3.5&29.7\\
& Gemma 3 27B IT& 7.7&71.6& 1.7&19.0\\
 & Qwen 3 30B IT& 3.8& 79.5&  0.0&16.5\\
 & Qwen 3 235B IT& 0.0& 85.1&  0.0&14.9\\
 & EDUMATH 12B (Ours)& 2.7&74.7& 9.3&13.3\\
& EDUMATH 30B (Ours)& 5.3&78.9& 0.0&15.8\\
\bottomrule
\end{tabular}
}
\caption{Classification of standards alignment errors for each model. See Figure \ref{fig:EDUMATH_standards_alignment_directions} for directions presented to our Gemma 3 27B IT automated annotator.}
\label{tab:stand_align_classification_EDUMATH}
\end{table*}

\section{Error Analysis} \label{error_analysis}
\paragraph{Common Error Types by Model}
Using our Gemma 3 27B IT annotator's CoT, we can also compute the frequencies of the types of errors each model makes to highlight their relative strengths/weaknesses. We can look at the proportion of overall MaC errors based on solvability, accuracy, educational appropriateness, and standards alignment as well as the proportion of each specific error type made for each criteria individually based on the response options given to human annotators reported in Appendix \ref{EDUMATH:annotator_directions} (with the exception of solvability, which only has one response option). These frequencies are calculated by prompting Gemma 3 27B IT to identify the specific criteria failure and error type for that criteria in the Gemma 3 27B IT annotator's CoT (see Appendix \ref{EDUMATH:error_type_annotation_prompt} for the prompt used for this experiment). This experiment thus allows us to pinpoint the specific error that makes each MWP not MaC, which is a framework other related work has followed \cite{ariyarathne_elementary_2025, christ_mathwell_2024, jiao_automatic_2023}. We show results for this experiment in Tables \ref{tab:errors_by_criteria_EDUMATH}, \ref{tab:accuracy_classification_EDUMATH}, \ref{tab:appropriateness_classification_EDUMATH}, and \ref{tab:stand_align_classification_EDUMATH}. 

Failures in any of the four criteria are equally harmful to student learning because a failure in any area makes the question inappropriate to give to a student. As shown in Table \ref{tab:errors_by_criteria_EDUMATH}, overall models tend to avoid making solvability and educational appropriateness errors. The majority of errors for each model, therefore, are for accuracy and standards alignment, which are the more nuanced criteria. 

As shown in the annotator directions in Appendix \ref{EDUMATH:annotator_directions}, there are two primary reasons a question would be labeled as inaccurate: 1) The solution is incorrect or 2) The solution is correct but contains a reasoning error, does not make sense, or is too complicated for a student/teacher to read, both of which would be equally harmful to student learning. As shown in Table \ref{tab:accuracy_classification_EDUMATH}, all models except GPT 4.5 make a majority of their accuracy errors from the latter error type, highlighting that they tend to arrive at the right answer most of the time, even when they make an accuracy error. 

As shown in Appendix \ref{EDUMATH:annotator_directions}, there are four primary reasons a MWP could be labeled as educationally inappropriate, including containing inappropriate material for a school setting, being strange or unrealistic, requiring no mathematical operations to solve, or being inappropriate for another reason. Of these errors, containing inappropriate material for a school setting would provide the most direct harm to young learners. As shown in Table \ref{tab:appropriateness_classification_EDUMATH}, no models make this most harmful error type, which is an improvement over the results reported in \citet{christ_mathwell_2024}
for older models.  EDUMATH 30B and the GPT models avoid educational appropriateness errors altogether, whereas the other models tend to make mistakes in outputting MWPs that are strange or unrealistic, with the exception that Gemma 3 27B IT has an equal share of MWPs that require no mathematical operations to solve. 

As shown in Appendix \ref{EDUMATH:annotator_directions}, there are four reasons why a MWP would not be aligned with a standard or standards: 1) It is too hard for the standard(s), 2) It does not address important parts of the standard(s),  3) It does not address the standard(s) at all, or 4) It requires the use of additional math topics beyond those listed in the standard(s), each of which would be equally harmful to student learning. Table \ref{tab:stand_align_classification_EDUMATH} shows the majority of standards alignment errors for each model are for MWPs that are missing important parts of the prespecified standard(s). Many models also frequently make MWPs that require additional math topics to solve, while it is less frequent that they make MWPs that are too hard to solve for the given standard(s) or do not address the standard(s) at all. These results suggest that models are generally able to incorporate some important elements from prespecified standard(s) when writing MWPs even when they do not fully align the problems with those standards. 

\section{Customizing MWPs to Student Interests} \label{customizing_mwps}
While \citet{christ_mathwell_2024} showed it was trivial to generate MWPs aligned with student interests with LLMs, we ran an experiment to verify this for our EDUMATH models. For this experiment, we generated 1,000 MWPs from both EDUMATH models based on a randomly selected topic from a list of 188 topics 3rd-5th grade students might be interested in. We then instructed Gemma 3 27B IT to indicate whether the MWP effectively incorporated the prespecified topic, which we used to calculate the proportion of generated MWPs that successfully included a topic (see Appendix \ref{EDUMATH:topic_annotation_prompt} for details on the prompts and topics used in this experiment). Based on this experiment, 95.5\% of EDUMATH 12B and 97.7\% of EDUMATH 30B MWPs successfully included a pre-specified topic, indicating both models are effectively able to incorporate student interests. 

\section{Student Study Details}\label{student_details}
\subsection{Student Study Full Results}\label{student_results}
Table \ref{tab:student_results} displays the main results from the student study for each experimental condition, which was approved by the institutional review board at our host university. As shown in the table, students in both experimental conditions solved the human-written and LLM-generated MWPs at a similar rate, with the differences in solve rate being statistically insignificant. However, students in both conditions preferred the LLM-generated MWPs to the human-written MWPs, though the preference for the LLM-generated MWPs was much stronger in School 2 (customization at the individual level), which is expected. In School 2, every student except one preferred the LLM-generated MWP over the human-written one. Table \ref{tab:student_preferences} shows a breakdown of the reasons students indicated for preferring one problem over another in each experimental condition. As shown in the table, in School 1 students who preferred the LLM-generated MWP most often preferred it because they liked the topic it was about (e.g., the student interest used to guide the generation), whereas their preference reason was more varied in cases where they preferred the human-written MWP. In School 2, the vast majority of students preferred the LLM-generated MWP because they liked the topic. In this condition, the one student who preferred the human-written MWP over the LLM-generated MWP preferred it because they liked the topic it was about. 
\begin{table*}[t]
    \centering
     \resizebox{\linewidth}{!}{
    \begin{tabular}{lccc}
    \toprule
        \textbf{Variable}& \textbf{\shortstack{Human-written MWP\\ Solve Rate}}&\textbf{\shortstack{LLM-generated MWP\\ Solve Rate}}&\textbf{\shortstack{Preferred\\ LLM-generated MWP}}\\
        \hline
        School 1 (customization at class level)& 68.2& 70.6&64.0*\\
        School 2 (customization at individual level)& 73.7& 68.4&94.7*\\ \bottomrule
    \end{tabular}}
    \caption{Main results for student study. A * denotes the difference in the share of students who preferred the LLM-generated MWP to the human-written MWP is statistically significant at the p < 0.01 level.}
    \label{tab:student_results}
\end{table*}
\begin{table*}[h]
    \centering
    \resizebox{\linewidth}{!}{
    \begin{tabular}{clcc}
        \toprule
         & \textbf{Preference Reason} & \textbf{School 1 (customization at class level)} & \textbf{School 2 (customization at individual level)} \\
        \midrule
        \parbox[t]{2mm}{\multirow{7}{*}{\rotatebox[origin=c]{90}{LLM-generated}}} 
         &Liked Topic & 48.5& 77.8\\
         & Easy& 17.6& 16.7\\
         & Challenging & 13.9& 0.0\\
         & Liked Math Operation & 10.3& 5.5\\
         & More Realistic& 0.0& 0.0\\
         & Contained More Context& 3.6& 0.0\\
         & Other& 6.1& 0.0\\
        \midrule
        \parbox[t]{2mm}{\multirow{7}{*}{\rotatebox[origin=c]{90}{Human-written}}} 
         & Liked Topic & 20.4& 100.0\\
         & Easy& 25.8& 0.0\\
         & Challenging & 19.4& 0.0\\
         & Liked Math Operation & 12.9& 0.0\\
         & More Realistic& 3.2& 0.0\\
         & Contained More Context& 0.0& 0.0\\
         & Other& 18.3& 0.0\\
        \bottomrule
    \end{tabular}}
    \caption{Breakdown of reasons indicated by students when explaining why they preferred one MWP over the other for each experimental condition.}
    \label{tab:student_preferences}
\end{table*}
\subsection{Student Demographics} \label{student_demographics}
Table \ref{tab:student_demographics} displays the school-level demographics for both schools included in the student study, while Table \ref{tab:condition2_breakdown} shows the number of students per grade level and number of worksheets per student in the second student experimental condition (customization at the individual level). We report demographics at the school level to avoid collecting any personally identifying information for individual students. While the first school is considerably more diverse than the second, all students included from the second school were receiving support from a math interventionist, suggesting they were struggling in school. 
\begin{table*}[t]
    \centering
     \resizebox{\linewidth}{!}{
    \begin{tabular}{lcc}
    \toprule
        \textbf{Variable}& \textbf{School 1 (customization at class level)}&\textbf{School 2 (customization at individual level)}\\
        \hline
        Economically Disadvantaged& 27.4&1.4\\
        English Language Learner& 18.3&6.9\\
        Race or Ethnicity&  &\\
        \hspace{3mm} White& 45.8&62.8\\
        \hspace{3mm} Black& 10.7&5.5\\
        \hspace{3mm} Asian& 12.2&15.2\\
        \hspace{3mm} Hispanic& 22.5&12.4\\
 \hspace{3mm} American Indian& 0.2&0.0\\
 \hspace{3mm} Native Hawaiian& 0.2&0.0\\
        \hspace{3mm} Mixed& 8.4&4.1\\ \bottomrule
    \end{tabular}}
    \caption{Demographics for students in both experimental conditions reported at the school level.}
    \label{tab:student_demographics}
\end{table*}
\begin{table}[t]
    \centering
    \resizebox{\linewidth}{!}{
    \begin{tabular}{lcc}
    \toprule
        \textbf{Grade Level}& \textbf{Number of Students}& \textbf{Number of Worksheets/Student}\\
        \toprule
        3& 2& 1\\
        4& 7& 2\\
        5& 3& 1\\
        \bottomrule
    \end{tabular}}
    \caption{Number of students and number of worksheets per student by grade level for the second experimental condition in the student study (customization at the individual level).}
    \label{tab:condition2_breakdown}
\end{table}
\subsection{Sample Student Worksheet} \label{sample_worksheet}
\begin{figure*}
    \centering
    \fbox{\includegraphics[width=1\linewidth]{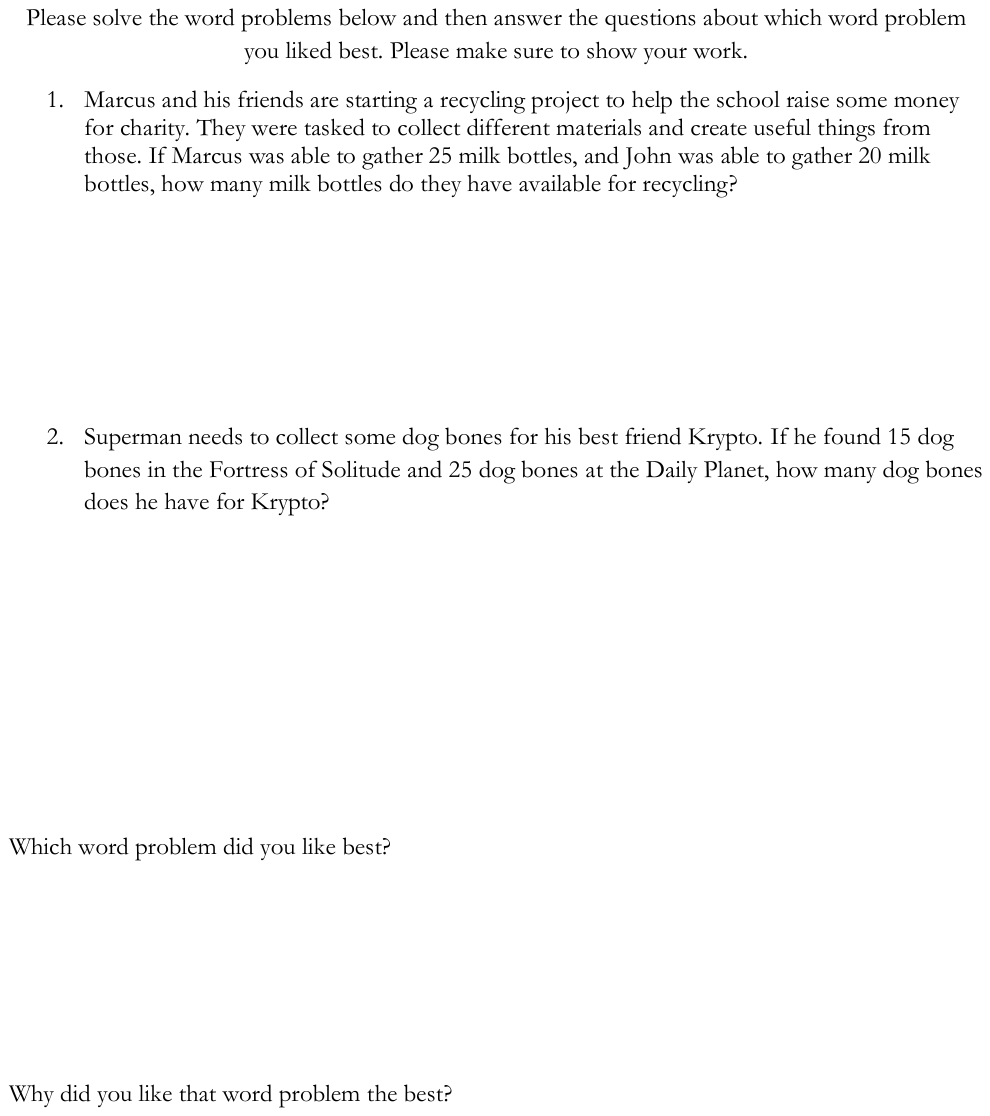}}
    \caption{Sample student worksheet for first experimental condition in the student study. The second MWP is the one customized to a student interest at the class level (Superman).}
    \label{fig:sample_student_worksheet}
\end{figure*}

Figure \ref{fig:sample_student_worksheet} shows an example student worksheet used for the first week of data collection in the first experimental condition (student interest customization at the class level rather than individual level).

\section{Sample Annotated Data} \label{EDUMATH:sample_annotated_data}
\subsection{Example that Meets All Criteria (MaC)}
\noindent\textbf{Grade Level:} 4 \newline

\noindent\textbf{Standard:}	Single-step or multi-step multiplication with whole numbers using a) a two-digit factor and a one-digit factor, b) a three-digit factor and a one-digit factor or c) a two-digit factor and a two-digit factor \newline

\noindent\textbf{Question:}  A Pokémon trainer has 14 boxes of Poké Balls to store her Pokémon. Each box can hold 25 Poké Balls. How many Poké Balls can the trainer store in total? \newline

\noindent\textbf{Solution:} 	
\newline To find the total number of Poké Balls the trainer can store, we need to multiply the number of boxes (14) by the number of Poké Balls each box can hold (25). \newline
14 x 25 = 350 \newline
This means the trainer can store a total of 350 Poké Balls. \newline
The final answer is 350.\newline

\noindent\textbf{Annotation:} This MWP MaC because it is solvable, has an accurate solution, contains content appropriate for a school setting, and is aligned with the pre-specified educational math standard. 

\subsection{Solvability Negative Example}
\noindent\textbf{Question:} Maria is baking cookies for her school bake sale. She plans to make batches of cookies with 2 1/6 cups of flour each. If she has 3 1/6 cups of flour, how much more flour does she need to buy to make all the cookie batches? \newline

\noindent\textbf{Annotation:} The problem does not define how many batches of cookies Maria wants to make, making it unsolvable. 

\subsection{Accuracy Negative Example}
\noindent\textbf{Question:} Ms. Carter has 378 pencils that she wants to divide equally among 14 classrooms. Each classroom should get the same number of pencils. How many pencils will be left after distributing them equally? \newline

\noindent\textbf{Solution:}
\newline To find out how many pencils will be left, we need to divide the total number of pencils (378) by the number of classrooms (14).  \newline
378 ÷ 14 = 27 with a remainder.  \newline
Let’s calculate it step by step:  \newline
14 × 27 = 378  \newline
But let's check how much is exactly divisible:  \newline
14 × 27 = 378  \newline
Since 14 × 27 = 378, that means all 378 pencils are used up.  \newline
Wait — let's divide correctly:  \newline
378 ÷ 14 = 27, because 14 × 27 = 378.  \newline
So, 378 - (14 × 27) = 378 - 378 = 0.  \newline
But let's double-check the multiplication:  \newline
14 × 20 = 280  \newline
14 × 7 = 98  \newline
280 + 98 = 378  \newline
Yes, 14 × 27 = 378  \newline
So, 378 ÷ 14 = 27 with no remainder.  \newline
But the problem asks how many are left — so if all are distributed, the remainder is 0.  \newline
Wait — let's do the division carefully:  \newline
378 ÷ 14  \newline
14 goes into 37 two times (14 × 2 = 28), subtract: 37 - 28 = 9  \newline
Bring down the 8 → 98  \newline
14 goes into 98 seven times (14 × 7 = 98), subtract: 98 - 98 = 0  \newline
So, 378 ÷ 14 = 27 with a remainder of 0.  \newline
Therefore, there are 0 pencils left.  \newline
But let's check: 14 × 27 = 378 — yes, exactly.  \newline
So, all pencils are distributed.  \newline
The final answer is 0.\newline

\noindent\textbf{Annotation:} This solution is inaccurate because it conducts many unnecessary checks to ensure the final answer is correct, leading it to be inappropriate to present to a student/teacher. 

\subsection{Educational Appropriateness Negative Example}
\noindent\textbf{Question:} Ms. Applebaum is organizing a field trip for her 4th-grade class. She has 25 permission slips to collect, but 7 students have already returned theirs. She wants to divide the remaining permission slips equally among the 13 students who still need to turn them in. What fraction of a permission slip does each student need to turn in? \newline

\noindent\textbf{Annotation:} This question is not educationally appropriate because it is not based in reality. It is not possible for students to turn in a fraction of a permission slip.

\subsection{Standards Alignment Negative Example}
\noindent\textbf{Grade Level:} 4 \newline

\noindent\textbf{Standard:}	Single-step or multi-step multiplication with whole numbers using a) a two-digit factor and a one-digit factor, b) a three-digit factor and a one-digit factor or c) a two-digit factor and a two-digit factor \newline

\noindent\textbf{Question:}  The school is preparing for the annual book fair. They ordered 24 boxes of books, and each box contains 15 books. If the school wants to arrange the books into displays with 8 books per display, how many displays can they make? \newline

\noindent\textbf{Solution:} 	
\newline First, we need to find the total number of books ordered. To do this, we multiply the number of boxes (24) by the number of books in each box (15). \newline
24 x 15 = 360 \newline
So, there are 360 books in total. \newline
Next, we need to determine how many displays they can make. To do this, we divide the total number of books (360) by the number of books per display (8). \newline
360 ÷ 8 = 45 \newline
Therefore, the school can make 45 displays. \newline
The final answer is 45. \newline

\noindent\textbf{Annotation:} This MWP is not standards aligned because the standard calls only for multiplication and the solution requires division in addition to multiplication.

\section{Example Problem from ASDIV Subset} \label{EDUMATH:asdiv_example}
Below is an example 4th grade MWP from the labeled ASDIV 3rd-5th grade subset. The question comes directly from the ASDIV dataset, while the solution is generated by Llama 3.3 70B IT using the prompt in Appendix \ref{EDUMATH:solution_generation_prompt}. \newline

\noindent \textbf{Example ASDIV 3rd-5th Grade Subset MWP}
\noindent \textbf{Grade:} 4 \newline

\noindent\textbf{Standard:} Single-step division with remainders for a one-digit divisor and a two- or three-digit dividend \newline

\noindent\textbf{Question:} Jawbreakers come in packages of 8. Sarah ate 20 Jawbreakers. How many Jawbreakers does she have left? \newline

\noindent\textbf{Solution:} \newline
To find out how many Jawbreakers Sarah has left, we first need to determine how many packages of Jawbreakers she bought. \newline
Since each package has 8 Jawbreakers, we need to divide the total number of Jawbreakers she ate (20) by the number of Jawbreakers in each package (8).\newline
20 / 8 = 2 with a remainder of 4 \newline
This means she ate 2 and a half packages of Jawbreakers. \newline
Since you cannot buy half a package, this means she must have bought at least 3 packages to eat 20 Jawbreakers (2 packages would only have 16 Jawbreakers). \newline
To find how many Jawbreakers she has left, we need to multiply the number of Jawbreakers per package (8) by the number of packages she bought (3) and subtract the total number she ate (20) \newline
8 x 3 = 24 \newline
24 - 20 = 4 \newline
The final answer is 4. 
\section{Prompts}
\subsection{Standards Annotation Prompt for ASDIV Subset} \label{EDUMATH:annotate_standards_prompt}
Figure \ref{fig:annotate_standards_prompt} displays the prompt for annotating the ASDIV 3rd-5th grade subset for standards and substandards using Llama 3.3 70B IT to ensure that each listed standard/substandard is incorporated in the problem, where the standards/substandards contain the direct text from the Virginia Standards of Learning (SOLs). We use a temperature of 0 to ensure the model selects the most probable next token during generation.
\begin{figure*}
  \centering
  \begin{tcolorbox}[
      colback=gray!5,
      colframe=black,
      width=0.9\linewidth,
      sharp corners,
      boxrule=0.6pt,
      left=4pt, right=4pt, top=4pt, bottom=4pt
    ]
    You are an experienced elementary school teacher. You are tasked with reading an educational standard(s) and its substandards and then assessing a word problem and its solution to determine whether the problem meets the standard(s) and substandards. When responding, say "Yes." or "No." to indicate whether the problem meets the specified standard(s). Please put detailed reasoning after "Yes." or "No." You only need to say "Yes." or "No." once to indicate whether the problem meets the standard(s) and substandards as a whole. Only say "Yes." if the problem exactly matches the operations and constraints mentioned in the standard(s) and substandards. For example, if the standard mentions multiplication within a certain number range, then the problem will only meet that standard if it requires multiplication within that range. \\[0.5em]
Standards: \{insert standards\} \\[0.5em]
Substandards: \{insert substandards\} \\[0.5em]
Word Problem: \{insert word problem\} \\[0.5em]
Solution: \{insert solution\} 
  \end{tcolorbox}
  \caption{Prompt for annotating ASDIV 3rd-5th grade subset for standards.}
  \label{fig:annotate_standards_prompt}
\end{figure*}
\subsection{Solution Generation Prompt for ASDIV Subset} \label{EDUMATH:solution_generation_prompt}
Figure \ref{fig:solution_generation_prompt} displays the prompt for generating step-by-step, readable solutions for the ASDIV 3rd-5th grade subset using Llama 3.3 70B IT. The few-shot examples are consistent across prompts and were manually constructed by our research team member with K-12 teaching experience. We use a temperature of 0 to ensure the model selects the most probable next token during generation.
\begin{figure*}
  \centering
  \begin{tcolorbox}[
      colback=gray!5,
      colframe=black,
      width=0.9\linewidth,
      sharp corners,
      boxrule=0.6pt,
      left=4pt, right=4pt, top=4pt, bottom=4pt,
    ]
    You are an experienced elementary school teacher. You are tasked with developing step-by-step solutions to math word problems for your students. The solutions should outline all the necessary steps, show complete work, and be written in a way a grade school student would understand. Make sure you separate your solution by writing ``Solution:'' and then the solution and end your solution with saying ``The final answer is \_'' where ``\_'' is filled in with the final answer. Here are some examples:\\[0.5em]
    \{3-shot examples\}\\[0.5em]
    Question: \{insert question\}\\[0.5em]
    Solution:
  \end{tcolorbox}
  \caption{Prompt for generating readable solutions to the ASDIV 3rd--5th grade subset.}
  \label{fig:solution_generation_prompt}
\end{figure*}
\subsection{Solution Annotation and Revision Prompts for ASDIV Subset} \label{EDUMATH:solution_revision_prompt}
Figures \ref{fig:solution_annotation_prompt} and \ref{fig:solution_revision_prompt} display the prompts for annotating the appropriateness of the solutions for the ASDIV 3rd-5th grade subset generated using the prompt in Appendix \ref{EDUMATH:solution_generation_prompt} and revising inappropriate solutions, respectively. Both prompts are passed to Gemma 3 27B IT and the few-shot examples are consistent across prompts and were manually constructed by our research team member with K-12 teaching experience. We use a temperature of 0 for both prompts to ensure the model selects the most probable next token during generation.
\begin{figure*}
  \centering
  \begin{tcolorbox}[
      colback=gray!5,
      colframe=black,
      width=0.9\linewidth,
      sharp corners,
      boxrule=0.6pt,
      left=4pt, right=4pt, top=4pt, bottom=4pt,
    ]
    You are an experienced elementary school teacher tasked with evaluating solutions for word problems for 3rd-5th grade students written by a less experienced teacher. The word problem and solution you will evaluate will be based on a grade level and math topic(s) and your job is to determine whether the solution is accurate, high quality, and appropriate/readable for young learners. \par\vspace{0.5em}

An accurate solution is one where the final answer and intermediate reasoning are both correct. If the final answer is correct but the intermediate reasoning is wrong, does not make sense, is unnecessarily repetitive, and/or is too complicated for a student/teacher to read, the solution is not accurate. \par\vspace{0.5em}

Here are some examples of successful evaluations: \par\vspace{0.5em}
\{7-shot examples\} \par\vspace{0.5em}

Now evaluate this word problem's solution and remember to answer "Yes." or "No." followed by your reasoning: \par\vspace{0.5em}
Grade Level: \{insert grade level\} \\[0.5em]
Math Topics: \{insert math topics\} \\[0.5em]
Question: \{insert question\} \\[0.5em]
Solution: \\[0.5em]
\{insert solution\} \\[0.5em]
Is the solution accurate?
  \end{tcolorbox}
  \caption{Prompt for annotating whether generated solutions for the ASDIV 3rd--5th grade subset are appropriate.}
  \label{fig:solution_annotation_prompt}
\end{figure*}
\begin{figure*}
  \centering
  \begin{tcolorbox}[
      colback=gray!5,
      colframe=black,
      width=0.9\linewidth,
      sharp corners,
      boxrule=0.6pt,
      left=4pt, right=4pt, top=4pt, bottom=4pt,
    ]
    You are an experienced elementary school teacher tasked with rewriting solutions for word problems for 3rd-5th grade students written by a less experienced teacher. The word problem and original solution will be based on a grade level and math topic(s) and will contain reasoning identifying what is wrong with the original solution. Your job is to use the reasoning to rewrite the solution to ensure it is accurate, high quality, and appropriate/readable for young learners. \\[0.5em]

An accurate solution is one where the final answer and intermediate reasoning are both correct, there is not unnecessary repetition, and the wording is not too complicated for a student/teacher to read.\\[0.5em]

Here are some examples of successful solution rewrites:\\[0.5em]
\{5-shot examples\} \\[0.5em]

Now rewrite this word problem's solution:\\[0.5em]
Grade Level: \{insert grade level\} \\[0.5em]
Math Topics: \{insert math topics\} \\[0.5em]
Question: \{insert question\} \\[0.5em]
Solution: \\[0.5em]
\{insert solution\} \\[0.5em]
Reasoning for why the solution is inaccurate: \\[0.5em]
\{insert reasoning\} \\[0.5em]
Corrected Solution:
  \end{tcolorbox}
  \caption{Prompt for revising inappropriate solutions for the ASDIV 3rd--5th grade subset as flagged using the prompt in Figure \ref{fig:solution_annotation_prompt}.}
  \label{fig:solution_revision_prompt}
\end{figure*}
\subsection{MWP Generation Prompt} \label{EDUMATH:mwp_generation_prompt}
Figure \ref{fig:mwp_generation_prompt} displays the standard prompt used for generating MWPs throughout this study. The 8-shot examples are filled in with randomly-selected examples for the particular grade level and standard for the query. STEM is used for the few-shot examples in the experiments in Section \ref{EDUMATH_experiments}, while the ASDIV 3rd-5th grade subset is used for the few-shot examples when generating annotation data in Section \ref{STEM_annotation}. For standards that were not in the ASDIV 3rd-5th grade subset when generating annotation data, the few-shot examples were populated with 8 randomly-selected samples from the same grade level as the standard. For all models, we use a temperature of 1 to strike a balance between creativity and selecting the most probable next token when generating MWPs. 

When customizing MWPs to student interests in Section \ref{customizing_mwps}, the formatted string "about \{topic\}" is added to the prompt such that the first sentence after the 8-shot examples reads, "Write a new question about \{topic\} based on the grade level and math topic(s) below." In this string, "\{topic\}" is filled in with a random selection from the list of topics reported in Appendix \ref{EDUMATH:topic_annotation_prompt}. 
\begin{figure*}
  \centering
  \begin{tcolorbox}[
      colback=gray!5,
      colframe=black,
      width=0.9\linewidth,
      sharp corners,
      boxrule=0.6pt,
      left=4pt, right=4pt, top=4pt, bottom=4pt,
    ]
    You are an experienced teacher tasked with writing word problems and solutions for 3rd-5th grade students. The question you write will be based on a grade level and math topic(s). The question's content should exactly match and incorporate ALL of the mathematical topics and constraints listed in the math topic(s). The question and answer pair you write should be solvable with the information presented in the question, contain an accurate solution, and contain language and context appropriate for a 3rd-5th grade student in a school setting (i.e., no harmful language and topics should be appropriate for school settings). \\[0.5em]
        
Here are some examples: \\[0.5em]
\{8-shot examples\} \\[0.5em]
     
Write a new question based on the grade level and math topic(s) below. Make sure to incorporate all of the information in the math topic(s) into your question; for example, if the math topic(s) mention remainders, the problem you write should require a remainder. Also make sure to write "Question:" followed by your question and "Solution:" followed by your solution. \\[0.5em]
Grade Level: \{insert grade level\} \\[0.5em]
Math Topics: \{insert math topics\} \\[0.5em]
Question:
  \end{tcolorbox}
  \caption{Standard prompt for generating new MWPs.}
  \label{fig:mwp_generation_prompt}
\end{figure*}
\subsection{Gemma 3 27B IT Annotator Prompt and Sample Output} \label{EDUMATH:automatic_annotation_prompt}
Figure \ref{fig:automatic_annotation_prompt} displays the prompt for annotating MWPs using Gemma 3 27B IT. The 25-shot examples are consistent across all model queries and are examples our research team manually constructed when aligning the automated annotator with expert feedback over multiple rounds of prompt iteration. We use a temperature of 0 to ensure the model selects the most probable next token during generation to improve accuracy and reproducibility of model annotations. Below is an example output for a MWP that did not MaC according to the annotator. \newline

\noindent \textbf{Example Gemma 3 27B IT Annotator Output}
\noindent \textbf{Grade:} 3 \newline

\noindent \textbf{Math Topic:}\newline
\noindent Single-step or multi-step addition and/or subtraction with whole numbers less than 1,000\newline

\noindent \textbf{Question:} Maria is making bracelets for her friends. She has 235 beads. She wants to make 6 bracelets and use the same number of beads on each bracelet. How many beads will be left over after she makes the bracelets? \newline

\noindent \textbf{Solution:} \newline
\noindent Maria has 235 beads and wants to make 6 bracelets.\newline

\noindent To find out how many beads will be left over, we need to divide the total number of beads (235) by the number of bracelets (6).\newline

\noindent 235 ÷ 6 = 39 with a remainder of 1.\newline

\noindent This means Maria can put 39 beads on each bracelet and she will have 1 bead left over.\newline

\noindent The final answer is 1. \newline

\noindent\textbf{Annotation:} No. While the question is solvable, has an accurate solution, and is educationally appropriate, it is not standards aligned. Specifically, the question only incorporates division and does not include addition or subtraction as stated in the math topic.
\onecolumn
\begin{center}
  \begin{tcolorbox}[
      colback=gray!5,
      colframe=black,
      width=0.9\linewidth,
      sharp corners,
      boxrule=0.6pt,
      left=4pt, right=4pt, top=4pt, bottom=4pt,
      breakable
    ]
    You are an experienced elementary school teacher tasked with evaluating word problems and solutions for 3rd-5th grade students written by a less experienced teacher. The word problem and solution you will evaluate will be based on a grade level and math topic(s) and your job is to determine whether the problem and solution are high quality.  \\[0.5em]
    
There are four criteria you will use to evaluate the word problem: solvability, accuracy, educational appropriateness, and standards alignment. Questions that meet all four criteria are labeled as high quality. Any word problem or solution that does not meet one or more of the criteria is labeled as not high quality. Here is more information about how to evaluate a word problem and solution based on the four criteria: \\[0.5em]
    
Solvability: \\[0.5em]
A solvable question means that it can be solved with the information present and does not contain a mathematical scenario that is impossible (e.g., giving away more money or items than you have).  \\[0.5em]
    
Accuracy: \\[0.5em]
An accurate solution is one where the final answer and intermediate reasoning are both correct. If the final answer is correct but the intermediate reasoning is wrong, does not make sense, is unnecessarily repetitive, and/or is too complicated for a student/teacher to read, the solution is not accurate.  \\[0.5em]
    
Educational Appropriateness: \\[0.5em]
An educationally appropriate question is one you would feel comfortable giving to a student in a 3rd-5th grade school setting. Educationally appropriate questions contain content and context appropriate for students in a school setting. There are four main reasons why a question would be educationally inappropriate:\\[0.5em]
1. It contains material inappropriate for a school setting (e.g., language about harming someone)\\[0.5em]
2. It is strange, confusing, contains conflicting information, and/or is not based in reality (e.g., contains misinformation)\\[0.5em]
3. It requires no mathematical operations to solve because it gives the answer away\\[0.5em]
4. It is inappropriate for a different reason \\[0.5em]
    
Standards alignment: \\[0.5em]
A standards aligned question is one that adequately addresses important elements from EACH pre-specified numbered math topic. If more than one math topic is listed, then the question should incorporate important elements of EACH numbered math topic. If only one math topic is included, then it is okay if the question only incorporates elements of that topic. You should only evaluate whether the question incorporates elements from EACH listed topic; you should not penalize questions that could incorporate other topics that are not in the numbered list of topics. If a specific numbered topic lists multiple mathematical operations like addition, subtraction, and/or division, it is okay if the problem just addresses one of those operations; if a topic says "the question may include OTHER TOPIC," then it is okay if the question does not include that other topic, as it is optional. However, the problem should incorporate meaningful elements of EACH numbered topic; for example, if a numbered topic lists decimal division, the problem should incorporate decimal division. There are four main reasons why a question would not be standards aligned: \\[0.5em]
1. It is too hard for the given topic(s)\\[0.5em]
2. It does not address some important parts of the numbered topic(s) or one or more of the numbered topic(s)\\[0.5em]
3. It does not address the numbered topic(s) at all\\[0.5em]
4. It requires additional math topics or operations that are not listed in the specified math topic(s)  \\[0.5em]

Here are some examples of successful evaluations: \\[0.5em]
\{25-shot examples\} \\[0.5em]
    
Now evaluate this word problem and remember to answer "Yes." or "No." followed by your reasoning. \\[0.5em]
Grade Level: \{insert grade level\} \\[0.5em]
Math Topics: \{insert math topics\} \\[0.5em]
Question: \{insert question\} \\[0.5em]
Solution: \\[0.5em]
\{insert solution\} \\[0.5em]
Is this question high quality?
  \end{tcolorbox}
  \captionof{figure}{Prompt for annotating MWPs with Gemma 3 27B IT.}
  \label{fig:automatic_annotation_prompt}
\end{center}
\twocolumn
\subsection{Gemma 3 27B IT Prompt for Identifying Failure Types in Annotated MWPs} \label{EDUMATH:error_type_annotation_prompt}
Figure \ref{fig:error_type_annotation_prompt} displays the prompt for labeling the error types in model-annotated MWPs using Gemma 3 27B IT. The 8-shot examples are consistent across model queries and were manually constructed by our research team. We use a temperature of 0 to ensure the model selects the most probable next token during generation.
\onecolumn
\begin{center}
  \begin{tcolorbox}[
      colback=gray!5,
      colframe=black,
      width=0.9\linewidth,
      sharp corners,
      boxrule=0.6pt,
      left=4pt, right=4pt, top=4pt, bottom=4pt,
      breakable
    ]
    You are an experienced elementary school teacher tasked with reading an evaluation of a word problem and solution for 3rd-5th grade students written by a less experienced teacher and identifying the main error in the problem and/or solution that they noticed. You will have acess to the grade level, math topic, question, solution, and the less experienced teacher's evaluation of the word problem. Your job is to read the evaluation and identify which of the four criteria below it failed to incorporate as well as which specific error it made for the criteria it failed to incorporate. If the word problem violates more than 1 criteria according to the evaluation, your job is to identify the most important criteria the word problem failed to incorporate. The criteria and associated errors are:\\[0.5em]
    
Solvability:\\[0.5em]
A solvable question means that it can be solved with the information present and does not contain a mathematical scenario that is impossible (e.g., giving away more money or items than you have). A question is not solvable for one of two reasons:\\[0.5em]
1. It cannot be solved with the information present\\[0.5em]
2. It contains a mathematical scenario that is impossible\\[0.5em]
    
Accuracy:\\[0.5em]
An accurate solution is one where the final answer and intermediate reasoning are both correct. A solution is inaccurate if:\\[0.5em]
1. The final answer is wrong\\[0.5em]
2. The final answer is correct but the intermediate reasoning is wrong, does not make sense, is unnecessarily repetitive, and/or is too complicated for a student/teacher to read\\[0.5em]
    
Educational Appropriateness:\\[0.5em]
An educationally appropriate question is one you would feel comfortable giving to a student in a 3rd-5th grade school setting. Educationally appropriate questions contain content and context appropriate for students in a school setting. There are four main reasons why a question would be educationally inappropriate:\\[0.5em]
1. It contains material inappropriate for a school setting (e.g., language about harming someone)\\[0.5em]
2. It is strange, confusing, contains conflicting information, and/or is not based in reality (e.g., contains misinformation)\\[0.5em]
3. It requires no mathematical operations to solve because it gives the answer away\\[0.5em]
4. It is inappropriate for a different reason\\[0.5em]
    
Standards alignment:\\[0.5em]
A standards aligned question is one that adequately addresses important elements from EACH pre-specified numbered math topic. There are four main reasons why a question would not be standards aligned:\\[0.5em]
1. It is too hard for the given topic(s)\\[0.5em]
2. It does not address some important parts of the numbered topic(s) or one or more of the numbered topic(s)\\[0.5em]
3. It does not address the numbered topic(s) at all\\[0.5em]
4. It requires additional math topics or operations that are not listed in the specified math topic(s) \\[0.5em]

Here are some examples of successful evaluations:\\[0.5em]
\{8-shot examples\}\\[0.5em]
    
Now evaluate this word problem and remember to both identify which criteria the problem failed to incorporate using the "Error Type:" heading and the specific error it made for that criteria using the "Specific Error:" heading. The text in the "Specific Error:" heading should exactly match one of the error types for the criteria defined above. Do not include any information other than the specific error type in the "Specific Error:" heading. Make sure to include BOTH the error type and the specific error in your response. \\[0.5em]
Grade Level: \{insert grade level\} \\[0.5em]
Math Topics: \{insert math topics\} \\[0.5em]
Question: \{insert question\} \\[0.5em]
Solution: \\[0.5em]
\{insert solution\} \\[0.5em]
Evaluation: \{insert model annotation\} \\[0.5em]
Error Type:
  \end{tcolorbox}
  \captionof{figure}{Prompt for labeling error types in model-annotated MWPs with Gemma 3 27B IT.}
  \label{fig:error_type_annotation_prompt}
\end{center}
\twocolumn
\subsection{Gemma 3 27B IT Student Interest Annotation Prompt} \label{EDUMATH:topic_annotation_prompt}
Figure \ref{fig:topic_annotation_prompt} displays the Gemma 3 27B IT prompt for annotating whether a generated MWP successfully incorporates a randomly-selected student interest from the list below. We use a temperature of 0 to ensure the model selects the most probable next token during generation. The list of student interests contains the 43 topics introduced in \citet{christ_mathwell_2024} as well as 145 additional topics obtained by prompting GPT 4o to generate a list of topics 3rd-5th grade students would be interested in to see whether our models are able to incorporate a diverse range of student interests into generated MWPs. 

\noindent\textbf{Topics:} ['Superman',
 'Batman',
 'Wonder Woman',
 'Barbie',
 'Power Rangers',
 'basketball',
 'soccer',
 'football',
 'volleyball',
 'field hockey',
 'Fortnite',
 'Spiderman',
 'Iron Man',
 'Captain America',
 'Captain Marvel',
 'Thor, the God of Thunder',
 'Black Panther',
 'Taylor Swift',
 'swimming',
 'Pokémon',
 'Super Mario',
 'Naruto',
 'unicorns',
 'Hello Kitty',
 'Minecraft',
 'lacrosse',
 'cheer leading',
 'LeBron James',
 'Steph Curry',
 'Patrick Mahomes',
 'Serena Williams',
 'dogs',
 'cats',
 'dinosaurs',
 'Harry Potter',
 'cars',
 'planes',
 'trains',
 'pizza',
 'cookies',
 'ice cream',
 'candy',
 'Frozen (Elsa and Anna)',
 'Star Wars',
 'Paw Patrol',
 'My Little Pony',
 'Minions',
 'Jurassic Park',
 'SpongeBob SquarePants',
 'Disney Princesses',
 'Toy Story',
 'The Incredibles',
 'Scooby-Doo',
 'Peppa Pig',
 'Dora the Explorer',
 'Pikachu',
 'Thomas the Tank Engine',
 'Sonic the Hedgehog',
 'Transformers',
 'Cinderella',
 'Moana',
 'Shrek',
 'Winnie the Pooh',
 'Tom and Jerry',
 'Sesame Street',
 'The Lion King',
 'Alice in Wonderland',
 'The Little Mermaid',
 'Peter Pan',
 'Aladdin',
 'The Jungle Book',
 'Pocahontas',
 'Beauty and the Beast',
 'Frozen',
 'Ratatouille',
 'Finding Nemo',
 'Cars',
 'Up',
 'The Simpsons',
 'Looney Tunes',
 'Teenage Mutant Ninja Turtles',
 'Mythical Creatures (dragons, unicorns)',
 'Dinosaurs',
 'Space and Astronauts',
 'Robots',
 'Aliens',
 'Exploring the Ocean',
 'Underwater Creatures',
 'Pirates',
 'Fairies',
 'Wizards',
 'Magic Tricks',
 'Time Travel',
 'Detectives and Mystery',
 'Inventions',
 'The Avengers',
 'The Justice League',
 'Dance and Ballet',
 'Music Instruments',
 'Art and Drawing',
 'Science Experiments',
 'Cooking and Baking',
 'DIY Crafts',
 'Board Games',
 'Puzzles',
 'Riddles',
 'Pets (cats, dogs, hamsters)',
 'Farm Animals',
 'Zoo Animals',
 'Wildlife Conservation',
 'Plants and Gardening',
 'Hiking and Nature',
 'Weather and Meteorology',
 'The Solar System',
 'Camping',
 'National Parks',
 'Trains and Railroads',
 'Planes and Aviation',
 'Cars and Racing',
 'Construction Vehicles',
 'Firefighters',
 'Police Officers',
 'Doctors and Nurses',
 'Astronauts and Space Exploration',
 'Animals and Wildlife',
 'Space and Astronomy',
 'Robots and Technology',
 'Underwater Life',
 'Fairy Tales and Folklore',
 'Outer Space',
 'Music and Instruments',
 'Insects and Bugs',
 'Historical Figures',
 'Countries and Cultures',
 'Mythical Creatures',
 'Magic and Wizards',
 'Friendship and Relationships',
 'Ocean Life',
 'Cars and Vehicles',
 'Famous Inventors',
 'Famous Artists',
 'Ancient Civilizations',
 'Space Exploration',
 'Gardening',
 'Environmental Conservation',
 'Pirates and Treasure',
 'Famous Scientists',
 'Computer Programming',
 'Unexplained Mysteries',
 'Planets and the Solar System',
 'Cartoons and Animated Shows',
 'Photography',
 'Books and Reading',
 'Volcanoes',
 'Mythology',
 'Ancient Egypt',
 'Reptiles and Amphibians',
 'Recycling',
 'Fairy Gardens',
 'Indoor Games',
 'Marine Biology',
 'Virtual Reality',
 'Natural Disasters',
 'Construction and Building',
 'the Circus and Performing Arts',
 'Science Fiction',
 'Pottery and Ceramics',
 'Famous Explorers',
 'Birds and Bird Watching',
 'Famous Landmarks',
 'Health and Nutrition',
 'Myths and Legends',
 'Fashion and Clothing',
 'DIY Science Projects',
 'Cultural Festivals',
 'Forests and Trees',
 'Mummies',
 'Famous Composers',
 'Circus Animals',
 'Geology',
 'Farm Life',
 'Travel and Adventure',
 'Ballet and Dance',
 'Whales and Dolphins',
 'Mystery Stories',
 'Hiking and Camping',
 'Games and Puzzles',
 'Space Aliens and UFOs']
\begin{figure*}
  \centering
  \begin{tcolorbox}[
      colback=gray!5,
      colframe=black,
      width=0.9\linewidth,
      sharp corners,
      boxrule=0.6pt,
      left=4pt, right=4pt, top=4pt, bottom=4pt,
    ]
    Your goal is to read a math word problem and determine whether it effectively includes a pre-specified topic. If the question does not effectively include the topic, write "No." followed by your reasoning. If the question does effectively include the topic, write "Yes." followed by your reasoning.\\[0.5em]
    
Now evaluate whether this problem includes the specified topic and remember to exactly answer "Yes." or "No." followed by your reasoning.\\[0.5em]
Topic: \{insert topic\} \\[0.5em]
Question: \{insert question\} \\[0.5em]
Does the question effectively incorporate the specified topic?
  \end{tcolorbox}
  \caption{Prompt for annotating whether MWPs successfully include a pre-specified topic.}
  \label{fig:topic_annotation_prompt}
\end{figure*}

\end{document}